\documentclass[10pt,twocolumn,letterpaper]{article}

\usepackage[pagenumbers]{wacv} 

\usepackage{graphicx}
\usepackage{amsmath}
\usepackage{amssymb}
\usepackage{booktabs}

\usepackage{bbding}
\usepackage{pifont}
\usepackage{wasysym}
\usepackage{multirow}

\usepackage[pagebackref,breaklinks,colorlinks]{hyperref}

\usepackage[capitalize]{cleveref}
\crefname{section}{Sec.}{Secs.}
\Crefname{section}{Section}{Sections}
\Crefname{table}{Table}{Tables}
\crefname{table}{Tab.}{Tabs.}

\def\wacvPaperID{818} 
\def\confName{WACV}
\def\confYear{2025}

\newcommand\blfootnote[1]{%
    \begingroup 
    \renewcommand\thefootnote{}\footnote{#1}%
    \addtocounter{footnote}{-1}%
    \endgroup 
}

\begin{document}

\title{MegaFusion: Extend Diffusion Models towards Higher-resolution Image Generation without Further Tuning}

\author{Haoning Wu$^{*}$, Shaocheng Shen$^{*}$, Qiang Hu$^{\dagger}$, Xiaoyun Zhang$^{\dagger}$, Ya Zhang, Yanfeng Wang\\[3pt]
Shanghai Jiao Tong University, China
}

\twocolumn[{%
\renewcommand\twocolumn[1][]{#1}%
\maketitle
\vspace{-1.0cm}
\begin{center}
   \centering
   \includegraphics[width=\textwidth]{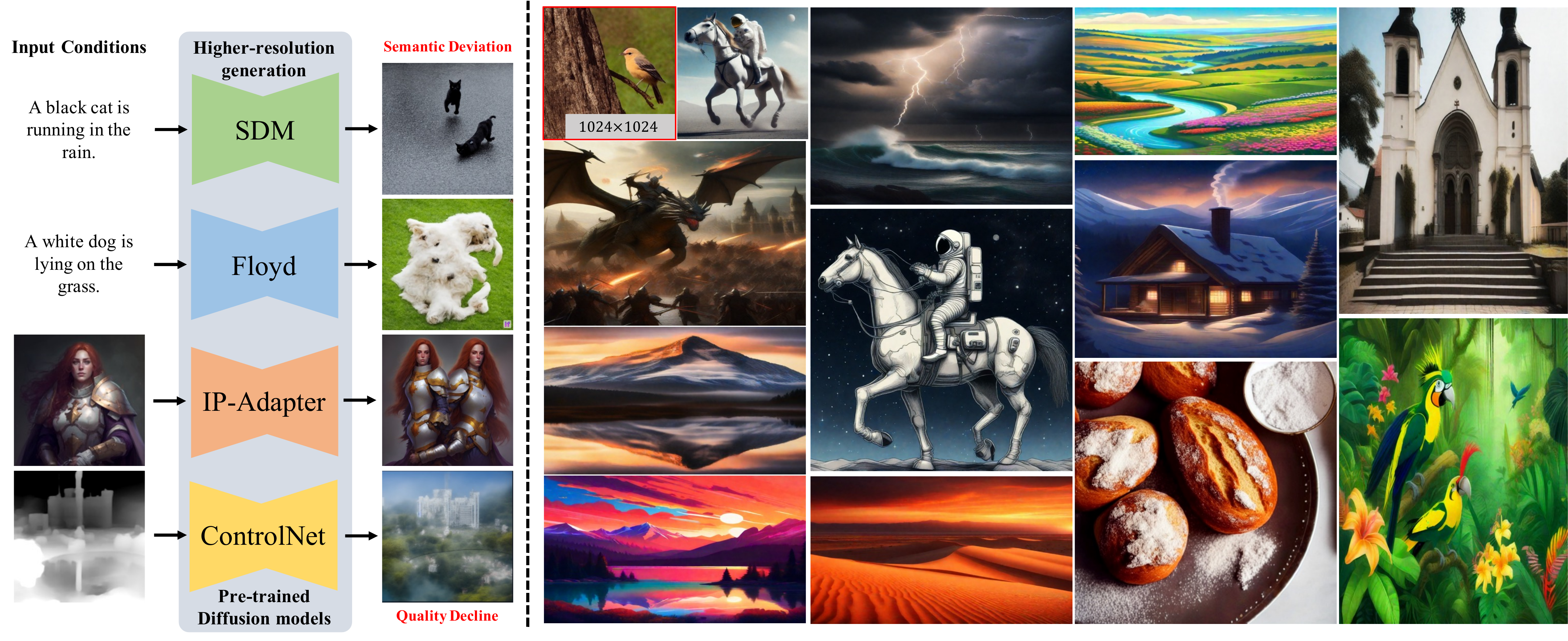} 
   \vspace{-0.7cm}
   \captionof{figure}{
      \textbf{Overview.} 
      {\em Left}: Existing diffusion-based text-to-image models fall short in synthesizing higher-resolution images due to the fixed image resolution during training, resulting in a noticeable decline in image quality and semantic deviation. 
      {\em Right:} Our proposed tuning-free \textbf{MegaFusion} can effectively and efficiently extend diffusion models (e.g. SDM~\cite{SDM}, SDXL~\cite{podell2023sdxl} and Floyd~\cite{DeepFloyd}) towards generating images at higher resolutions (e.g., $1024 \times 1024, 1920 \times 1080, 2048 \times 1536,$ and $2048 \times 2048$) of arbitrary aspect ratios (e.g., $1:1, 16:9,$ and $4:3$).
      We recommend the reader to zoom in for the visualization results.
      }
  \label{fig:teaser}
 \end{center}
 }]

 \blfootnote{
 \hspace{-0.575cm} *: These authors contribute equally to this work. \\ 
$\dagger$: Corresponding author.
}


\begin{abstract}

Diffusion models have emerged as frontrunners in text-to-image generation, but their fixed image resolution during training often leads to challenges in high-resolution image generation, such as semantic deviations and object replication. 
This paper introduces \textbf{MegaFusion}, a novel approach that extends existing diffusion-based text-to-image models towards efficient higher-resolution generation without additional fine-tuning or adaptation.
Specifically, we employ an innovative \textbf{{\em truncate and relay}} strategy to bridge the denoising processes across different resolutions, allowing for high-resolution image generation in a coarse-to-fine manner.
Moreover, by integrating {\em dilated convolutions} and {\em noise re-scheduling}, we further adapt the model's priors for higher resolution.
The versatility and efficacy of MegaFusion make it universally applicable to both latent-space and pixel-space diffusion models, along with other derivative models. 
Extensive experiments confirm that MegaFusion significantly boosts the capability of existing models to produce images of megapixels and various aspect ratios, while only requiring about $\mathbf{40\%}$ of the original computational cost.
Code is available at \url{https://haoningwu3639.github.io/MegaFusion/}.

\end{abstract}
\section{Introduction}
Diffusion models have demonstrated unparalleled performance across broad applications such as text-to-image generation~\cite{ho2020denoising, song2020denoising, ho2022classifier, ho2022imagen, SDM, DeepFloyd, esserSD3}, image editing~\cite{brooks2022instructpix2pix, meng2021sdedit, kawar2022imagic, hertz2022prompt, xie2022smartbrush, lugmayr2022repaint, chenpixart}, consistent image sequence generation~\cite{liu2024intelligent, maharana2022storydall, pan2022synthesizing}, and even achieves promising results in challenging text-to-video generation~\cite{ho2022video, li2023videogen, ho2022imagen, singer2022make}.
Among them, Stable Diffusion (also known as Latent Diffusion~\cite{SDM}) performs denoising in a compressed latent space, and has showcased impressive generative capabilities after pre-training on large-scale paired datasets~\cite{schuhmann2022laionb}.
In comparison, Imagen~\cite{saharia2022photorealistic} and Floyd~\cite{DeepFloyd} adopt cascading diffusion models in pixel space, initiating with low-resolution image synthesis followed by successive super-resolution stages.

Despite these advancements, as depicted in Figure~\ref{fig:teaser} {\em Left}, these models face a major challenge: they struggle to generate images beyond training resolutions, leading to semantic deviation and degraded image quality.
Existing solutions often require additional tuning or are limited to specific models.
For example, MultiDiffusion~\cite{multidiffusion} and ElasticDiffusion~\cite{haji2023elasticdiffusion} adopt post-processing optimization to stitch high-resolution panoramas, which is inefficient and time-consuming.
Relay Diffusion~\cite{teng2024relay} employs blurring diffusion in pixel space, yet it necessitates training multiple specific diffusion models from scratch.
ResAdapter~\cite{cheng2024resadapter} and CheapScaling~\cite{guo2024make} involve minimal extra training through LoRA~\cite{hu2021lora} or Upsamplers, but still incur a notable training overhead.
Meanwhile, tuning-free alternatives like ScaleCrafter~\cite{he2023scalecrafter} and FouriScale~\cite{huang2024fouriscale} adapt pre-trained SDMs for higher resolutions, but demand meticulous hyperparameter adjustment and are restricted to latent-space models.

To tackle these limitations, we introduce \textbf{MegaFusion}, a tuning-free method to extend existing diffusion models towards generating higher-resolution and variable aspect ratio images with megapixels.
Concretely, we begin with a {\em truncate and relay} strategy, which seamlessly bridges the synthesis of different resolution images, enabling efficient generation in a coarse-to-fine manner with only $40\%$ of the original computational cost.
Moreover, it is orthogonally compatible with existing techniques such as {\em dilated convolutions}~\cite{yu2016multiscale} and {\em noise re-scheduling} for better image quality.
The versatility of \textbf{MegaFusion} makes it applicable to both latent-space and pixel-space diffusion models, as well as other diffusion-based frameworks with extra conditions, such as IP-Adapter~\cite{ye2023ip-adapter} and ControlNet~\cite{controlnet}.
As shown in Figure~\ref{fig:teaser} {\em Right}, MegaFusion significantly improves the ability of diffusion models to synthesize higher-resolution images with accurate semantics and superior quality.

To summarize, our contributions are fourfold:
(i) we propose \textbf{MegaFusion}, a tuning-free approach utilizing a {\em truncate and relay} strategy to efficiently generate high-quality, high-resolution images with megapixels in a coarse-to-fine manner;
(ii) we incorporate {\em dilated convolution} and {\em noise re-scheduling} techniques to further refine the adaptability of pre-trained diffusion models for higher resolution;
(iii) we demonstrate the applicability of our method across both latent-space and pixel-space diffusion models, as well as their extensions, synthesizing high-resolution images with various aspect ratios at roughly 40\% of the original computational cost;
(iv) we conduct extensive experiments validating the superiority of our proposed method, in terms of efficiency, image quality, and semantic accuracy.
\section{Related Works}
\noindent{\textbf{Diffusion Models.}} 
As a part of probabilistic generative models, diffusion models typically learn to generate samples by iterative denoising.
DDPM~\cite{ho2020denoising} has first showcased remarkable performance, while DDIM~\cite{song2020denoising} significantly improves sampling efficiency.
Leveraging their excellent generative capabilities, diffusion models have been applied to diverse fields, including image-to-image translation~\cite{brooks2022instructpix2pix, meng2021sdedit, kawar2022imagic, hertz2022prompt, xie2022smartbrush} and video generation~\cite{ho2022video, li2023videogen, ho2022imagen, singer2022make}.

\vspace{2pt}
\noindent{\textbf{Text-to-Image Generation.}}
Generative models have been widely adopted for the challenging text-to-image generation task, with GAN~\cite{goodfellow2020generative, zhang2017stackgan, xu2018attngan} as the pioneers.
Meanwhile, auto-regressive transformers like DALL·E~\cite{ramesh2021zero} further push the boundaries.
Diffusion models, such as DALL·E 2~\cite{ramesh2022hierarchical}, Imagen~\cite{saharia2022photorealistic} and Floyd~\cite{DeepFloyd}, have recently risen to prominence.
Notably, Stable Diffusion (Latent Diffusion~\cite{SDM}), performing denoising in latent space, has demonstrated outstanding performance, thereby being widely applied within the research community.
Additionally, SDXL~\cite{podell2023sdxl} further elevates the generative performance of Stable Diffusion with a diffusion refiner and an extra text encoder.

Our \textbf{MegaFusion}, is designed for seamless integration with diffusion models across both latent and pixel spaces, extending their capacity for higher-resolution generation.

\vspace{2pt}
\noindent{\textbf{Higher-resolution Generation.}}
Existing diffusion models are typically limited to fixed resolutions and aspect ratios, struggling to produce images beyond their training resolutions.
MultiDiffusion~\cite{multidiffusion} and ElasticDiffusion~\cite{haji2023elasticdiffusion} address this by synthesizing overlapping crops and merging them into panoramic images, which requires a time-consuming inference procedure.
Relay Diffusion~\cite{teng2024relay} designs a pixel-space model with blurring diffusion to craft high-resolution images, at the cost of retraining several models from scratch.
ScaleCrafter~\cite{he2023scalecrafter} achieves high-resolution generation by enlarging receptive fields of Stable Diffusion with dispersed convolution without extra fine-tuning.
DemoFusion~\cite{du2024demofusion} attempts to connect multiple resolutions for coarse-to-fine generation, but demands repeating generation multiple times, leading to low efficiency.

Several concurrent works also express rich interest in this task: 
ResAdapter~\cite{cheng2024resadapter} and CheapScaling~\cite{guo2024make} can generate images with unrestricted resolutions and aspect ratios with minimal tuning via trainable LoRA adapters or Upsamplers.
FouriScale~\cite{huang2024fouriscale} and HiDiffusion~\cite{zhang2023hidiffusion} offer a training-free strategy but are still limited to SDM-based models.

In contrast to the aforementioned methods, which either require further training or are limited to specific models, our proposed \textbf{MegaFusion} emerges as a versatile tuning-free solution that can be integrated seamlessly into existing diffusion models, enabling the synthesis of higher-resolution images of various aspect ratios.


\section{Preliminary}
In this section, we briefly introduce diffusion models, including their forward and backward processes in Sec.~\ref{diffusion_models}; 
and the latent diffusion models (LDMs) that perform diffusion in latent space to improve efficiency in Sec.~\ref{latent_diffusion_models}.

\subsection{Diffusion Models}
\label{diffusion_models}
Diffusion models, a class of deep generative models, iteratively transform Gaussian noise into structured data samples through a denoising process.
Concretely, diffusion models comprise a forward diffusion process that progressively adds Gaussian noise to an image $\mathbf{x}_0$ via a Markov process over $T$ steps.
Let $\mathbf{x}_t$ represent the noisy image at step $t$, with the transition from $\mathbf{x}_{t-1}$ to $\mathbf{x}_t$ being modeled by 
$q(\mathbf{x}_t | \mathbf{x}_{t-1}) = \mathcal{N}(\mathbf{x}_t; \sqrt{1 - \beta_t}\mathbf{x}_{t-1}, \beta_t \mathbf{I})$. 
Here, $\beta_t \in (0, 1)$ are pre-determined hyperparameters controlling the variance introduced at each step.
By defining $\alpha_t = 1 - \beta_t$ and $\bar{\alpha}_t = \prod_{i=1}^{t} \alpha_i$, we can leverage the properties of Gaussian distributions and the reparameterization trick to reformulate the relationship as:
$q(\mathbf{x}_t | \mathbf{x}_0) = \mathcal{N}(\mathbf{x}_t; \sqrt{\bar{\alpha}_t}\mathbf{x}_0, (1 - \bar{\alpha}_t)\mathbf{I})$.
This insight allows us to succinctly express the forward process with Gaussian noise $\epsilon$ as: $\mathbf{x}_t = \sqrt{\bar{\alpha}_t}\mathbf{x}_0 + \sqrt{1 - \bar{\alpha}_t} \epsilon$.

Diffusion models also encompass a reverse diffusion process to reconstruct images from noise.
This process, denoted as $p_{\theta}$, usually leverages a UNet-based~\cite{ronneberger2015u} model to estimate the noise term $\epsilon_{\theta}$, represented as:
$p_{\theta}(\mathbf{x}_{t-1} | \mathbf{x}_t) = \mathcal{N}(\mathbf{x}_t; \mu_{\theta}(\mathbf{x}_t, t), \Sigma_{\theta}(\mathbf{x}_t, t))$.
Here, $\mu_{\theta}$ is the predicted mean of Gaussian distribution, expressed in terms of the estimated noise $\epsilon_{\theta}$ as:
$\mu_{\theta}(\mathbf{x}_t, t) = \frac{1}{\sqrt{\alpha_t}}(\mathbf{x}_t - \frac{1 - \alpha_t}{\sqrt{1 - \bar{\alpha}_t}}\epsilon_{\theta}(\mathbf{x}_t, t))$

\subsection{Latent Diffusion Models}
\label{latent_diffusion_models}
To improve efficiency and reduce computational cost, Latent Diffusion (LDMs) execute diffusion and denoising within a learned low-dimensional latent space of a pre-trained Variational Autoencoder (VAE).
Specifically, the VAE encoder $\mathcal{E}$ maps an image $\mathbf{x}_0 \in \mathbb{R}^{3\times H\times W}$ to a latent representation $\mathbf{z}_0 \in \mathbb{R}^{4\times h\times w}$ via $\mathbf{z}_0 = \mathcal{E}(\mathbf{x}_0)$.
Afterwards, the decoder $\mathcal{D}$ reconstructs the original image $\mathbf{x}_0$ from $\mathbf{z}_0$, represented as $\hat{\mathbf{x}}_0 = \mathcal{D}(\mathbf{z}_0)\approx \mathbf{x}_0$.

This setup allows the diffusion process to be conducted in a compact latent space, facilitating efficient image synthesis.
During inference, LDM samples latent codes from a conditional distribution $p(\mathbf{z}_0 | c)$, where $c$ represents the conditional information such as text embedding from CLIP~\cite{CLIP} or T5~\cite{raffel2020exploring} text encoder.
This process can be formalized as:
$p_{\theta}(\mathbf{z}_{t-1} | \mathbf{z}_t, c) = \mathcal{N}(\mathbf{z}_t; \mu_{\theta}(\mathbf{z}_t, t, c), \Sigma_{\theta}(\mathbf{z}_t, t, c))$.

\begin{figure*}[thb]
  \centering
  \includegraphics[width=\textwidth]{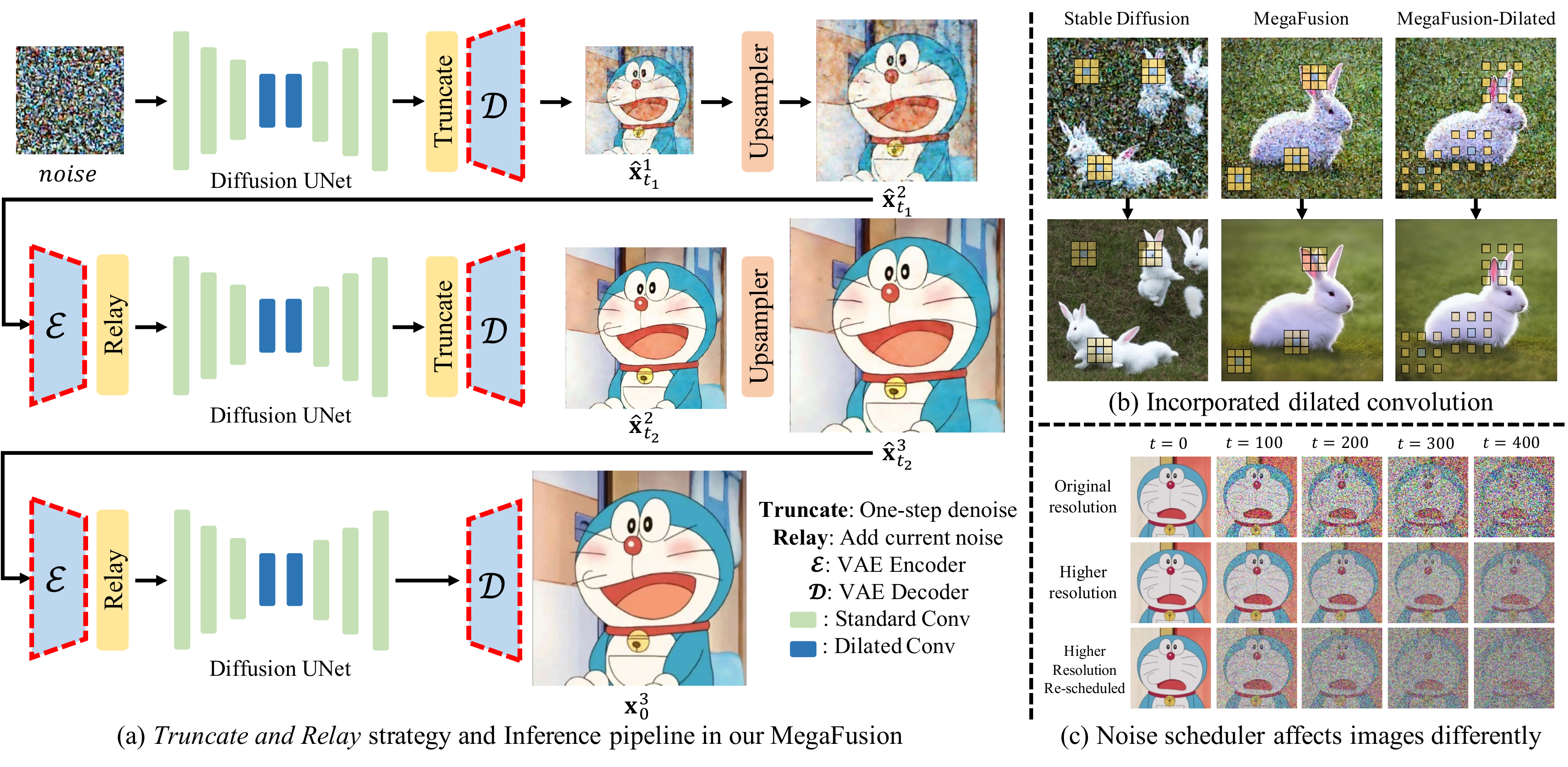} \\
  \vspace{-0.2cm}
  \caption{
  \textbf{Architecture Overview}.
  (a) The {\em Truncate and Relay} strategy in \textbf{MegaFusion} seamlessly connects generation processes across different resolutions to produce higher-resolution images without extra tuning, exemplified by a three-stage pipeline.
  For pixel-space models, the VAE encoder and decoder can be directly removed.
  (b) Limited receptive fields lead to quality decline and object replication.
  {\em Dilated convolutions} expand the receptive field at higher resolutions, enabling the model to capture more global information for more accurate semantics and image details.
  (c) Noise at identical timesteps affects images of different resolutions differently, deviating from the model's prior. 
  {\em Noise re-scheduling} helps align the noise level of higher-resolution images with that of the original resolution.
    }
    \vspace{-0.4cm}
 \label{fig:arch}
\end{figure*}

\section{Method}
This section initiates with elaborating on the {\em truncate and relay} strategy within our proposed tuning-free \textbf{MegaFusion} in Sec.~\ref{4.1}.
Then, we incorporate {\em dilated convolution} and {\em noise re-scheduling} to further adapt model prior towards higher resolution in Sec.~\ref{4.2}.
Lastly, we detail the application of our method across latent-space and pixel-space diffusion models, as well as their extensions, in Sec.~\ref{4.3}.

\subsection{Truncate and Relay Strategy}
\label{4.1}
\noindent{{\textbf{High-level Idea.}}
As evidenced by eDiff-I~\cite{balaji2022eDiff-I}, diffusion models synthesize semantics during early denoising steps and texture details in later steps.
Our intuition and insight here are that: we should perform early-stage denoising at original inference resolutions to guarantee accurate semantics, followed by {\em \textbf{truncating}} and {\em \textbf{relaying}} at higher resolutions to continue later-stage denoising to produce texture details. 
This strategy extends the higher-resolution generation capabilities of pre-trained models, enabling the synthesis of high-quality images with precise semantics at low computational costs, and supports various aspect ratios.

\vspace{2pt}
\noindent{\textbf{Problem Setting.}}
For clarity, we focus on latent-space diffusion models as an example.
As for pixel-space models, our method can be applied more straightforwardly and conveniently.
Given that our strategy is inherently tuning-free, we focus on the inference stage herein.
Using a pre-trained Latent Diffusion (LDM) with a denoiser $\epsilon_{\theta}$, 
a low-resolution latent code $\mathbf{z}_{0}^{1} \in \mathbb{R}^{4\times h_1\times w_1}$ can be synthesized within $T$ denoising steps conditioned on a text prompt $c_T$, and then decoded into an image $\mathbf{x}_{0}^{1} \in \mathbb{R}^{3 \times H_1 \times W_1}$ by the VAE decoder $\mathcal{D}$.
Our goal is to generate a higher-resolution image $\mathbf{x}_{0}^{k} \in \mathbb{R}^{3 \times H_k \times W_k}$ alongside its latent code $\mathbf{z}_{0}^{k} \in \mathbb{R}^{4\times h_k\times w_k}$, by linking generation processes across different resolutions over a total of $T$ steps, where $T = \sum_{i=1}^{k}T_i$.

\vspace{2pt}
\noindent{\textbf{Truncate.}}
To guarantee accurate semantics, we begin with a low-resolution generation through $T_1$ steps denoising:
\begin{align}
\label{equation:stage1}
    \mathbf{z}_{t-1} =  \frac{1}{\sqrt{\alpha_t}} & (\mathbf{z}_t - \frac{1 - \alpha_t}{\sqrt{1 - \bar{\alpha}_t}}\epsilon_{\theta}(\mathbf{z}_t, t, c_T)) + \sigma_t \epsilon, \\
    & \textrm{where} \,\,  t = T, T-1,..., T - T_1 + 1 \nonumber
\end{align}
where $\sigma_t$ are pre-calculated coefficients and $\epsilon$ denotes noise sampled from a standard Gaussian distribution.
Subsequently, at step $t_1 = T - T_1 + 1$, we {\em \textbf{truncate}} the generation process and compute the approximate clean latent code $\hat{\mathbf{z}}_{t_1}^{1} \in \mathbb{R}^{4 \times h_1 \times w_1}$, which serves as a pivotal element for multi-resolutions bridging via:
\begin{align}
\label{equation:denoise}
    \hat{\mathbf{z}}_{t_1}^{1} = \frac{1}{\sqrt{\bar{\alpha}_{t_1}}} {(\mathbf{z}_{t_1} - \sqrt{1 - \bar{\alpha}_{t_1}} \epsilon_{\theta} (\mathbf{z}_{t_1}, t_1, c_T))}
\end{align}
Here, $\hat{\mathbf{z}}_{t_1}^{1}$ is subsequently decoded to an image $\hat{\mathbf{x}}_{t_1}^{1} \in \mathbb{R}^{3 \times H_1 \times W_1}$ and upsampled to a higher-resolution relatively clean image $\hat{\mathbf{x}}_{t_1}^{2} \in \mathbb{R}^{3 \times H_2 \times W_2}$ utilizing a non-parametric Upsampler, $\Phi$, represented as:
\begin{align}
\label{equation:upsample}
    \hat{\mathbf{x}}_{t_1}^{1} = \mathcal{D}(\hat{\mathbf{z}}_{t_1}^{1}), \quad
    \hat{\mathbf{x}}_{t_1}^{2} = \Phi(\hat{\mathbf{x}}_{t_1}^{1})
\end{align}

\vspace{2pt}
\noindent{\textbf{Relay.}}
To further enhance higher-resolution texture details, the upsampled image $\hat{\mathbf{x}}_{t_1}^{2}$ is then re-encoded into to latent code $\hat{\mathbf{z}}_{t_1}^{2} \in \mathbb{R}^{4 \times h_2 \times w_2}$ via the VAE encoder $\mathcal{E}$ and perturbed with noise at the current step $t_1$ to {\em \textbf{relay}} the generation:
\begin{align}
\label{equation:addnoise}
    \hat{\mathbf{z}}_{t_1}^{2} = \mathcal{E}(\hat{\mathbf{x}}_{t_1}^{2}), \quad
    \mathbf{z}_{t_1}^{2} = \mathcal{N} (\mathbf{z}_{t_1}^{2}; \sqrt{\bar{\alpha}_{t_1}} \hat{\mathbf{z}}_{t_1}^{2}, (1 - \bar{\alpha}_{t_1}) \mathbf{I})
\end{align}
The generation process continues at a higher resolution, by re-leveraging Equation~\ref{equation:stage1} for $T_2$ steps of denoising, sequentially navigating through $t = T - T_1, T - T_1 - 1, ..., T - T_1 - T_2 + 1$.
Subsequently, the {\em truncate and relay} operations can be then conducted at step $t_2 = T - T_1 - T_2 + 1$.

As depicted in Figure~\ref{fig:arch} (a), this iterative process is repeated multiple times until the generation of a high-resolution latent code $\mathbf{z}_{0}^{k}$, which can be then decoded into a corresponding high-resolution image $\mathbf{x}_{0}^{k}$ with megapixels.

\begin{table*}[t]
\begin{center}
\tabcolsep=0.055cm
\small
\begin{tabular}{c|c|c|c|c|c|c|c|c|c|c|c}
\toprule
Methods & resolution & $\mathrm{FID}_r\downarrow$ & $\mathrm{FID}_b\downarrow$ & $\mathrm{KID}_r\downarrow$ & $\mathrm{KID}_b\downarrow$ & CLIP-T$\uparrow$ & CIDEr$\uparrow$ & Meteor$\uparrow$ & ROUGE$\uparrow$ & GFlops & Inference time \\ 
\midrule
SDM~\cite{SDM} & $1024 \times 1024$ & 41.35 & 51.02 & 0.0086 & 0.0113 & 0.3009 & 17.75 & 18.38 & 23.64 & 135.0K & 15.17s \\
SDM-StableSR~\cite{wang2023stablesr} & $1024 \times 1024$ & 25.46 & 19.61 & \textcolor{blue}{\underline{0.0062}} & \textcolor{blue}{\underline{0.0031}} & 0.3117 & 20.24 & 20.91 & 26.28 & 292.6K & 33.48s \\
SDM-RealESRGAN & $1024 \times 1024$ & \textcolor{blue}{\underline{25.20}} & 19.49 & \textcolor{red}{\textbf{0.0059}} & 0.0032 & 0.3119 & \textcolor{red}{\textbf{21.35}} & 21.26 & 26.76 & 35.6K & 5.12s \\
ResAdapter~\cite{cheng2024resadapter} & $1024 \times 1024$ & 27.38 & 20.47 & 0.0073 & 0.0033 & 0.3102 & 20.99 & 21.38 & 27.65 & 137.5K & 16.25s \\
ScaleCrafter~\cite{he2023scalecrafter} & $1024 \times 1024$ & 27.97 & 22.05 & 0.0076 & 0.0043 & \textcolor{red}{\textbf{0.3125}} & 20.14 & \textcolor{blue}{\underline{21.65}} & \textcolor{blue}{\underline{28.23}} & 135.0K & 17.52s \\ 
SDM-MegaFusion & $1024 \times 1024$ & 30.19 & \textcolor{blue}{\underline{10.98}} & 0.0088 & 0.0034 & 0.3101 & \textcolor{blue}{\underline{21.14}} & 21.44 & 27.34 & \textbf{48.2K} & \textbf{7.56s} \\
SDM-MegaFusion++ & $1024 \times 1024$ & \textcolor{red}{\textbf{25.14}} & \textcolor{red}{\textbf{7.82}} & 0.0064 & \textcolor{red}{\textbf{0.0012}} & \textcolor{blue}{\underline{0.3121}} & 20.46 & \textcolor{red}{\textbf{22.18}} & \textcolor{red}{\textbf{28.36}} & \textbf{48.2K} & \textbf{7.56s} \\
\midrule
SDXL~\cite{podell2023sdxl} & $2048 \times 2048$ & 47.53 & 47.08 & 0.0133 & 0.0139 & 0.3041 & 17.55 & 18.65 & 25.10 & 540.2K & 79.66s \\ 
SDXL-RealESRGAN & $2048 \times 2048$ & 24.76 & 13.54 & \textcolor{red}{\textbf{0.0056}} & \textcolor{blue}{\underline{0.0021}} & 0.3192 & 23.27 & 22.44 & \textcolor{blue}{\underline{28.44}} & 147.1K & 22.33s \\
ScaleCrafter~\cite{he2023scalecrafter} & $2048 \times 2048$ & 27.46 & 24.73 & 0.0064 & 0.0061 & 0.3138 & 19.97 & 22.34 & 28.12 & 540.2K & 80.72s \\ 
DemoFusion~\cite{du2024demofusion} & $2048 \times 2048$ & \textcolor{blue}{\underline{24.61}} & 13.36 & 0.0066 & 0.0023 & 0.3198 & 22.02 & \textcolor{red}{\textbf{22.86}} & \textcolor{red}{\textbf{28.48}} & 1354.9K & 217.19s \\ 
SDXL-MegaFusion & $2048 \times 2048$ & 25.12 &\textcolor{blue}{\underline{12.13}}  & \textcolor{blue}{\underline{0.0059}} & 0.0027 & \textcolor{blue}{\underline{0.3227}} & \textcolor{red}{\textbf{23.49}} & 22.65 & 28.12 & \textbf{216.1K} & \textbf{30.94s} \\ 
SDXL-MegaFusion++ & $2048 \times 2048$ & \textcolor{red}{\textbf{23.86}} &\textcolor{red}{\textbf{6.93}}  & \textcolor{red}{\textbf{0.0056}} &\textcolor{red}{\textbf{0.0018}}  & \textcolor{red}{\textbf{0.3244}} & \textcolor{blue}{\underline{23.42}} & \textcolor{blue}{\underline{22.74}} & 28.38 & \textbf{216.1K} & \textbf{30.94s} \\ 
\midrule
SD3~\cite{esserSD3} & $2048 \times 2048$ & \textcolor{blue}{\underline{38.37}} & \textcolor{blue}{\underline{31.91}} & \textcolor{blue}{\underline{0.0165}} & \textcolor{blue}{\underline{0.0181}} & \textcolor{blue}{\underline{0.3058}} & \textcolor{blue}{\underline{17.89}} & \textcolor{blue}{\underline{18.72}} & \textcolor{blue}{\underline{24.66}} & 433.9K & 64.89s \\
SD3-MegaFusion & $2048 \times 2048$ & \textcolor{red}{\textbf{28.81}} & \textcolor{red}{\textbf{8.93}} & \textcolor{red}{\textbf{0.0098}} & \textcolor{red}{\textbf{0.0018}} & \textcolor{red}{\textbf{0.3178}} & \textcolor{red}{\textbf{23.01}} & \textcolor{red}{\textbf{22.45}} & \textcolor{red}{\textbf{29.14}} & \textbf{201.4K} & \textbf{29.07s} \\
\midrule
Floyd-Stage1~\cite{DeepFloyd} & $128 \times 128$ & 66.27 & 81.65 & \textcolor{blue}{\underline{0.0262}} & 0.0454 & 0.2818 & 14.69 & 18.22 & 25.06 & 111.7K & 77.08s \\ 
Floyd-MegaFusion & $128 \times 128$ & \textcolor{blue}{\underline{53.09}} & \textcolor{red}{\textbf{39.73}} & 0.0273 & \textcolor{red}{\textbf{0.0334}} & \textcolor{blue}{\underline{0.3024}} & \textcolor{red}{\textbf{25.01}} & \textcolor{blue}{\underline{25.00}} & \textcolor{blue}{\underline{31.35}} & \textbf{44.9K} & \textbf{32.19s} \\
Floyd-MegaFusion++ & $128 \times 128$ & \textcolor{red}{\textbf{43.43}} & \textcolor{blue}{\underline{50.08}} & \textcolor{red}{\textbf{0.0213}} & \textcolor{blue}{\underline{0.0437}} & \textcolor{red}{\textbf{0.3046}} & \textcolor{blue}{\underline{20.28}} & \textcolor{red}{\textbf{25.01}} & \textcolor{red}{\textbf{31.64}} & \textbf{44.9K} & \textbf{32.19s} \\
\midrule
Floyd-Stage2~\cite{DeepFloyd} & $512 \times 512$ & 46.64 & 38.15 & 0.0254 & 0.0166 & 0.3098 & \textcolor{blue}{\underline{23.85}} & 21.47 & 26.26 & 60.7K & 48.58s  \\ 
Floyd-MegaFusion & $512 \times 512$ & \textcolor{blue}{\underline{39.80}} & \textcolor{blue}{\underline{24.87}} & \textcolor{blue}{\underline{0.0164}} &\textcolor{blue}{\underline{0.0078}} & \textcolor{blue}{\underline{0.3106}} & 23.22 & \textcolor{blue}{\underline{23.51}} & \textcolor{blue}{\underline{29.30}} & \textbf{24.3K} & \textbf{21.72s} \\ 
Floyd-MegaFusion++ & $512 \times 512$ & \textcolor{red}{\textbf{26.34}} & \textcolor{red}{\textbf{24.55}} & \textcolor{red}{\textbf{0.0063}} & \textcolor{red}{\textbf{0.0077}} & \textcolor{red}{\textbf{0.3110}} & \textcolor{red}{\textbf{24.01}} & \textcolor{red}{\textbf{23.58}} & \textcolor{red}{\textbf{29.52}} & \textbf{24.3K} & \textbf{21.72s} \\ 
\bottomrule
\end{tabular}
\end{center}
\vspace{-0.4cm}
\caption{\textbf{Quantitative comparison}. 
We compare our boosted models on higher-resolution generation with representative latent-space and pixel-space diffusion models on MS-COCO~\cite{lin2014microsoft} dataset.
\textcolor{red}{\textbf{RED}} represents best performance, and \textcolor{blue}{\underline{BLUE}} denotes second best performance.
}
\label{tab:quantitative_results}
\vspace{-0.4cm}
\end{table*}

\subsection{MegaFusion++}
\label{4.2}
Our MegaFusion, based on the {\em truncate and relay} strategy, can be further combined orthogonally with existing techniques such as {\em dilated convolution} and {\em noise rescheduling}, to adapt the model priors to higher resolutions.

\vspace{2pt}
\noindent{\textbf{Dilated Convolution.}}
Blurriness and semantic deviation in high-resolution images generated by diffusion models often stem from the constrained receptive field of UNet layers trained on fixed-resolution data, lacking comprehensive global context. 
As illustrated in Figure~\ref{fig:arch} (b), existing models trained on low-resolution images tend to synthesize multiple rabbits in different local regions due to insufficient receptive fields, leading to semantic inaccuracies. 
Inspired by ScaleCrafter~\cite{he2023scalecrafter}, we modify the convolutional kernels of the UNet-based denoiser $\epsilon_{\theta}$ to incorporate dilated convolutions~\cite{yu2016multiscale} with a specific dilation rate $\delta$. 
This broadens the receptive field without additional tuning, allowing for better incorporation of global information.

For the sake of simplicity, we omit channel dimensions and convolution biases here, focusing 
on modifying weight parameters to transform standard convolutions into dilated ones.
Given a feature map $F \in \mathbb{R}^{m\times n}$ and a convolutional kernel $k\in \mathbb{R}^{r \times r}$, the standard convolution can be represented as:
$(F * k)(p) = \sum_{s + t = p} F(s) \cdot k(t)$.
In contrast, the corresponding dilated convolution, with dilation rate $\delta$, can be expressed as:
$(F *_\delta k)(p) = \sum_{s + \delta t = p} F(s) \cdot k(t)$.
Here, $p, k,$ and $t$ denote spatial locations within the feature map and convolution kernel, respectively.

Following previous practices~\cite{he2023scalecrafter},
instead of replacing all convolutions with dilated ones, which may lead to catastrophic quality decline, we selectively apply this modification to the middle layers of UNet.
\textbf{The insight here is that}: we broaden receptive fields in the bottleneck to aggregate global context, while preserving original priors at higher resolution to sample nearby features for enhancing details.

\vspace{2pt}
\noindent{\textbf{Noise Re-scheduling.}}
Consistent with discoveries in simple diffusion~\cite{hoogeboom2023simple} and relay diffusion~\cite{teng2024relay}, we observe that identical noise levels impact images differently across various resolutions, as illustrated in Figure~\ref{fig:arch} (c), leading to varying signal-to-noise ratios (SNR) at the same timestep.

According to the SNR definition in previous work~\cite{hoogeboom2023simple}:
$SNR_t = \frac{(\sqrt{\bar{\alpha}}_t)^2}{(\sqrt{1 - \bar{\alpha}}_t)^2} = \frac{\bar{\alpha}_t}{1 - \bar{\alpha}_t}$.
Given a low-resolution image $\mathbf{x} \in \mathbb{R}^{3 \times H \times W}$ and a high-resolution one $\mathbf{x}' \in \mathbb{R}^{3 \times H' \times W'}$ 
with $H' > H$ and $W' > W$,
if we downsample $\mathbf{x}'$ to ${\mathbf{x}'}_{down} \in \mathbb{R}^{3 \times H \times W}$, the SNR at timestep $t$ of ${\mathbf{x}'}_{down}$ (denoted as ${SNR}_{down}^{H' \times W'}$) in comparison to $\mathbf{x}$ (represented as ${SNR}^{H\times W}$) will exhibit the following relationship:
${SNR}^{H \times W} = \gamma \cdot {SNR}_{down}^{H' \times W'}$.

Assuming the original noise scheduler at $H \times W$ is denoted as $\bar{\alpha}_t$, the revised scheduler ${\bar{\alpha}_t}'$ at higher resolution $H' \times W'$ should satisfy:
$\frac{\bar{\alpha}_t}{1 - \bar{\alpha}_t} = \gamma \cdot \frac{{\bar{\alpha}_t}'}{1 - {\bar{\alpha}_t}'}$.
This yields the relationship: ${\bar{\alpha}_t}' = \frac{{\bar{\alpha}_t}}{\gamma - (\gamma - 1) {\bar{\alpha}_t}}$.
Incorporating this into the high-resolution noise scheduler initialization gives a new ${\bar{\alpha}_t}$ sequence. 
This process, termed noise re-scheduling, adjusts noise levels to better suit higher-resolution image generation, thereby improving synthesis quality and fidelity.

\begin{figure*}[tph]
  \centering
  \includegraphics[width=\textwidth]{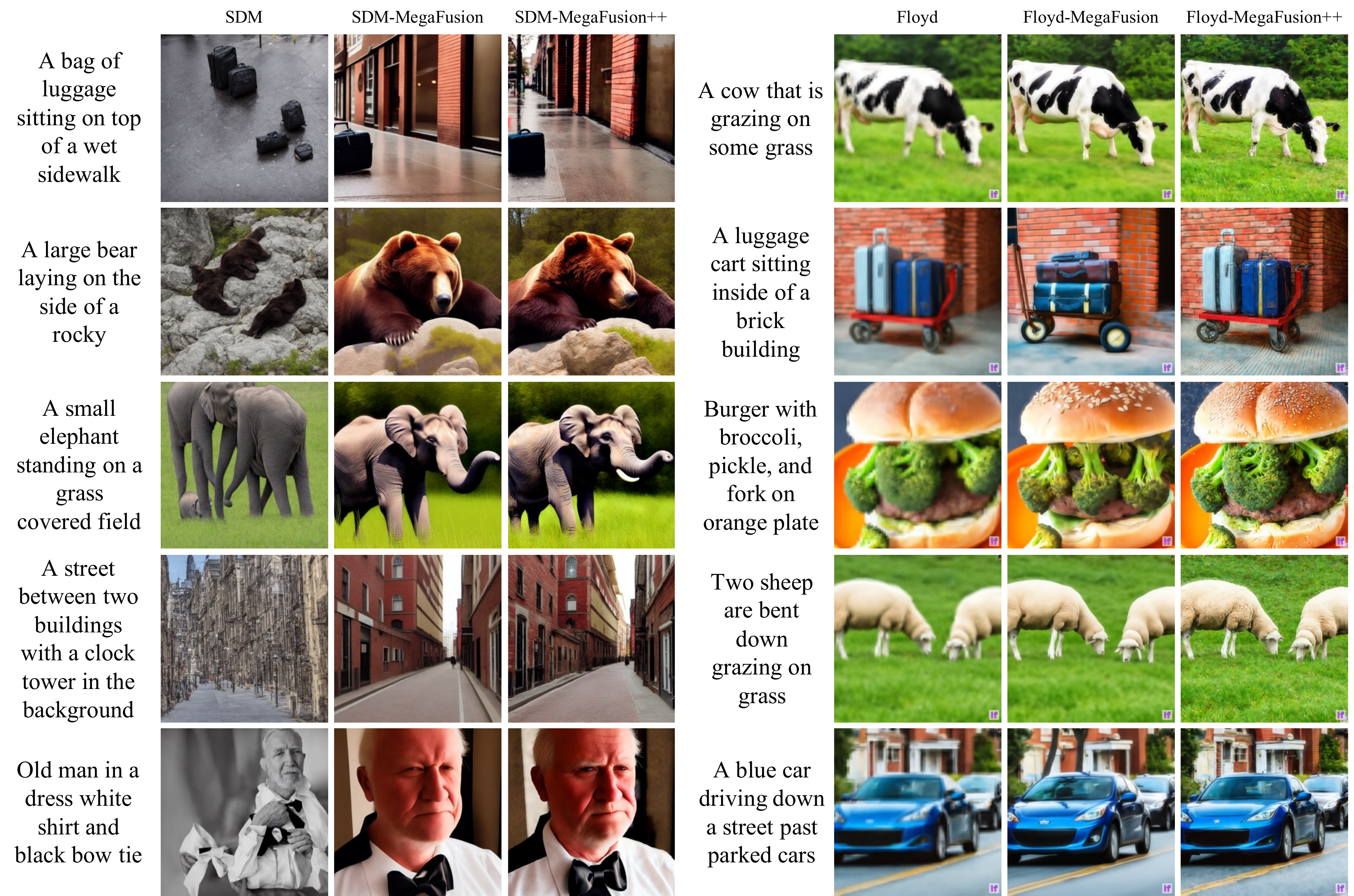} \\
  \vspace{-0.2cm} 
  \caption{
  \textbf{Qualitative results} of applying our MegaFusion to both latent-space and pixel-space diffusion models for higher-resolution image generation on MS-COCO and commonly used prompts from the Internet.
  Our method can effectively extend existing diffusion-based models towards synthesizing higher-resolution images of megapixels with correct semantics and details.
    }
    \vspace{-0.3cm}
 \label{fig:qualitative}
\end{figure*}

\subsection{Further Application on other Models}
\label{4.3}
\noindent{\textbf{Pixel-space Diffusion Models.}}
MegaFusion is equally applicable to pixel-space diffusion models, such as Floyd~\cite{DeepFloyd}, with the primary difference being that the {\em truncate and relay} operation is performed directly in pixel space. 
Consequently, Equations~\ref{equation:denoise}, \ref{equation:upsample}, and \ref{equation:addnoise} are adapted as follows:
\begin{align}
    \hat{\mathbf{x}}_{t_1}^{1} = \frac{1}{\sqrt{\bar{\alpha}_{t_1}}} {(\mathbf{x}_{t_1} - \sqrt{1 - \bar{\alpha}_{t_1}} \epsilon_{\theta} (\mathbf{x}_{t_1}, t_1, c_T))} \\
    \hat{\mathbf{x}}_{t_1}^{2} = \Phi(\hat{\mathbf{x}}_{t_1}^{1}), \,\,
    \mathbf{x}_{t_1}^{2} = \mathcal{N} (\mathbf{x}_{t_1}^{2}; \sqrt{\bar{\alpha}_{t_1}} \hat{\mathbf{x}}_{t_1}^{2}, (1 - \bar{\alpha}_{t_1}) \mathbf{I})
\end{align}
\noindent{\textbf{Diffusion Models with Extra Conditions.}}
Our methodology can also extend to diffusion models that incorporate extra input conditions, such as ControlNet~\cite{controlnet} and IP-Adapter~\cite{ye2023ip-adapter}. 
These models utilize both text condition $c_T$ and image condition $c_I$ as inputs. 
Consequently, Equation~\ref{equation:stage1} can be reformulated to accommodate both conditions:
\begin{align}
        \mathbf{z}_{t-1} = \frac{1}{\sqrt{\alpha_t}} (\mathbf{z}_t - \frac{1 - \alpha_t}{\sqrt{1 - \bar{\alpha}_t}}\epsilon(\mathbf{z}_t, t, c_T, c_I)) + \sigma_t \epsilon
\end{align}

\section{Experiments}
In this section, we first describe our experimental settings in Sec.~\ref{5.1}.
Next, we present comparisons to existing models with quantitative metrics and human evaluation in Sec.~\ref{5.2}.
We then showcase qualitative results of applying our method to various diffusion models in Sec.~\ref{5.3}.
Lastly, ablation studies are presented in Sec.~\ref{5.4}.

\subsection{Experiment Settings}
\label{5.1}
\noindent{\textbf{Implementation Details.}}
We evaluate text-to-image diffusion models in both latent space (SDM 1.5~\cite{SDM}, SDXL~\cite{podell2023sdxl}
and SD3~\cite{esserSD3})
and pixel space (Floyd~\cite{DeepFloyd}). 
All models use DDIM~\cite{song2020denoising} for $T=50$ steps of sampling unless explicitly stated otherwise.
Given that SDM is trained with a fixed resolution of $512 \times 512$, we choose to generate high-resolution images of $1024 \times 1024$ for quantitative comparison.
Specifically, we orchestrate generation across $k=3$ resolutions: $512$, $768$ and $1024$, with respective denoising steps of $T_1 = 40$, $T_2 = 5$, and $T_3 = 5$.
Furthermore, our proposed MegaFusion can be applied to synthesize even higher-resolution images with SDM for qualitative assessment.
SDXL defaults to synthesizing images of $1024 \times 1024$, considering the balance of computational costs, we discard the Refiner module and employ two-stage generation, $1024$ and $2048$, with their denoising steps being $T_1 = 40$ and $T_2 = 10$.
For SD3, which defaults to denoise 28 steps for generating $1024\times 1024$ images, we iterate $20$ steps at $1024$ resolution and $8$ steps at $2048$ resolution.

On the other hand, Floyd, a 3-stage cascaded model, sequentially upscales images from $64 \times 64$ to $256 \times 256$, culminating in $1024 \times 1024$ images.
Due to computational constraints, only the first two stages of Floyd are employed in our experiments.
The first stage necessitates $100$ sampling steps ($T_1 = 80$ for generating $64 \times 64$ images, and $T_2 = 20$ for $128 \times 128$), while the second stage requires $50$ steps ($T_1 = 40$ for $256 \times 256$ and $T_2 = 10$ for $512 \times 512$).

Bicubic upsampling serves as the default non-parametric Upsampler $\Phi$.
For typical $2\times$ higher-resolution generation, we set the dilation rate $\delta = 2$, and select the hyperparameter $\gamma = 4$ for noise re-scheduling.
For classifier-free guidance, to ensure a fair comparison, we apply the default weight $w$ of official implementations across all methods: $w=7.0$ for SDM, SDXL, SD3, and Floyd-Stage 1, and $w=4.0$ for Floyd-Stage 2.
All experiments are conducted on a single Nvidia RTX A40 GPU, with SDM, SDXL, 
and SD3
in $float16$ precision, and Floyd at in $float32$ precision.

\vspace{2pt}
\noindent{\textbf{Evaluation Datasets.}}
We assess our method and baseline models on the MS-COCO~\cite{lin2014microsoft} dataset, which comprises approximately 120K images in total, each accompanied by 5 captions.
Due to the computational costs of high-resolution generation, we randomly sample 10K images from MS-COCO, assigning a fixed caption to each as input.
To ensure consistent comparisons, we utilize the same random seed for each image across methods, neutralizing randomness.
For qualitative human evaluations, we use commonly available prompts from the Internet as text conditions and conditional images from the official code repositories as extra inputs for IP-Adapter and ControlNet.

\vspace{2pt}
\noindent{\textbf{Evaluation Metrics.}}
To evaluate the quality of generated images, we adopt several widely-used metrics, including Fréchet Inception Distance score (FID)~\cite{heusel2017gans}, Kernel Inception Distance score (KID), and CLIP~\cite{CLIP} text-image similarity (CLIP-T).
Following~\cite{he2023scalecrafter}, we consider two types of FID and KID: (i) $\mathrm{FID_r}$ and $\mathrm{KID_r}$ to gauge the quality and diversity of generated images relative to real ones, and (ii) $\mathrm{FID_b}$ and $\mathrm{KID_b}$ to assess the discrepancies between synthesized samples under the base training resolution and high resolution.
These latter metrics reflect the model's ability to retain generative proficiency at unfamiliar resolutions.

To evaluate the semantic accuracy of generated contents, we adopt MiniGPT-v2~\cite{chen2023minigptv2} to caption the images, and calculate several linguistic metrics between these captions and the original input text.
Concretely, we report the commonly used CIDEr~\cite{vedantam2015cider}, Meteor~\cite{banerjee2005meteor}, and ROUGE~\cite{lin2004rouge}.
Moreover, we detail the GFlops and inference time measured on a single A40 GPU for efficiency comparison.

\begin{figure*}[thp]
  \centering
  \includegraphics[width=.95\textwidth]{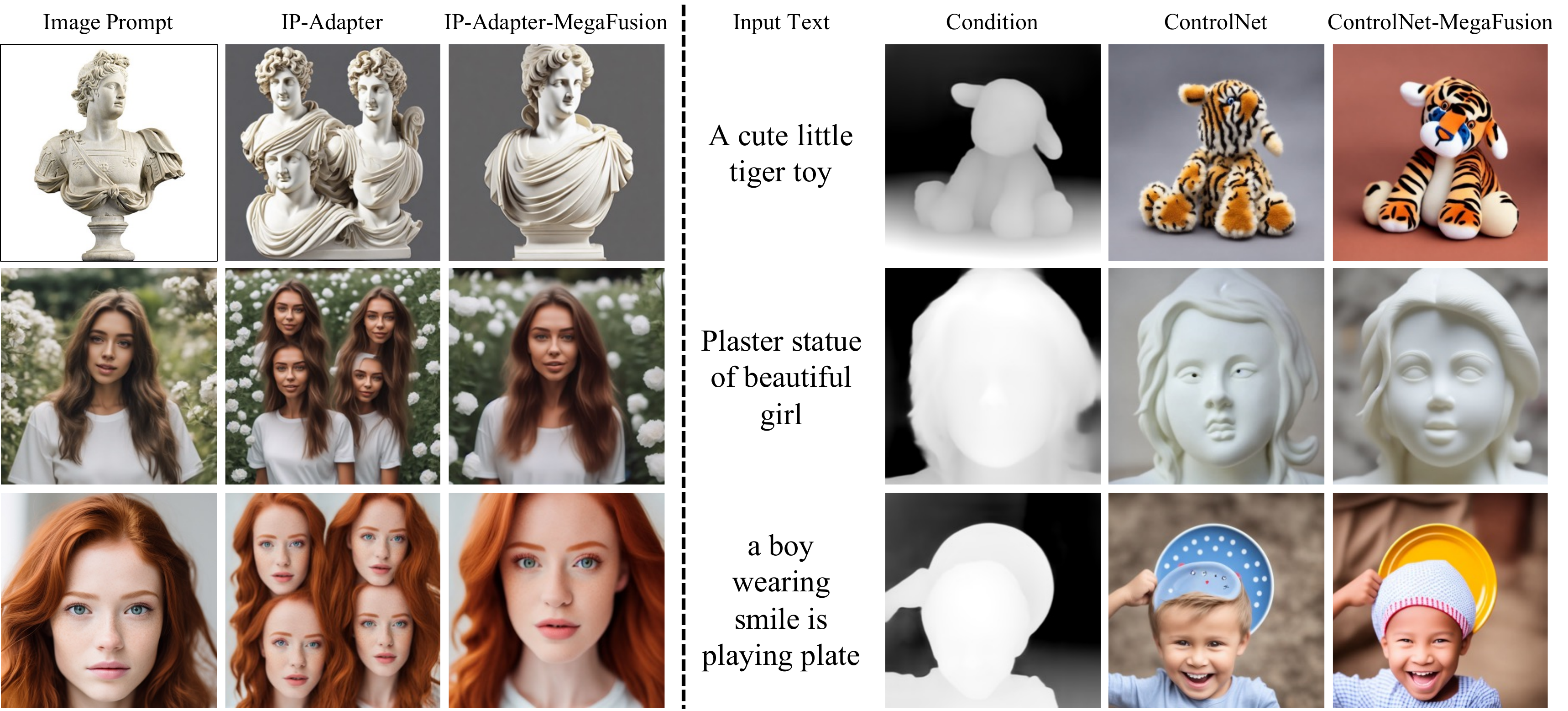} \\
  \vspace{-0.2cm}
  \caption{
  \textbf{Qualitative results} of incorporating MegaFusion to models with extra conditional inputs. 
  MegaFusion can be universally applied across various diffusion models, providing the capability for higher-resolution image generation with better semantics and fidelity.
}    
 \label{fig:qualitative2}
 \vspace{-0.5cm}
\end{figure*}

\subsection{Quantitative Results}
\label{5.2}
\noindent{\textbf{Objective Metrics.}}
We evaluate the performance of both latent-space and pixel-space models boosted by \textbf{MegaFusion} against their baseline counterparts on the MS-COCO~\cite{lin2014microsoft} dataset.
Here, [model-MegaFusion] refers to models employing {\em truncate and relay} strategy to bridge multi-resolution generation, while [model-MegaFusion++] denotes advanced models incorporating {\em dilated convolution} and {\em noise re-scheduling}.
We also compare several state-of-the-art methods, such as
ScaleCrafter~\cite{he2023scalecrafter}, and DemoFusion~\cite{du2024demofusion}, which are limited to specific latent-space models and less efficient, as well as SDM and SDXL with SR postprocessing~\cite{wang2023stablesr, LAR-SR, wang2021realesrgan, wang2018esrgan}, {\em e.g.} StableSR, and RealESRGAN.

The results in Table~\ref{tab:quantitative_results} highlight significant improvements with MegaFusion across all metrics, including image quality, semantic accuracy, and especially computational efficiency. 
This confirms that MegaFusion effectively extends the generative capabilities of existing diffusion models towards synthesizing high-resolution images with correct semantics and details at only 40\% of the original computational cost. 
Moreover, incorporating {\em dilated convolution} and {\em noise re-scheduling} further improves performance on several metrics, reflecting improved generation diversity and better alignment with real images and text conditions.

\vspace{2pt}
\noindent{\textbf{Human Assessment.}}
To complement our objective analysis, we conduct a human-centric evaluation focusing on image quality and semantic integrity. 
Concretely, utilizing identical text prompts and random seed, we synthesize higher-resolution images via standard models (SDM and Floyd) and their MegaFusion-boosted counterparts.
Participants are asked to rate the outputs with a score from 1 to 5 (higher is better), considering both image quality and semantic accuracy.
Additionally, they also need to select their preferred image among the options for preference rating.

The results in Table~\ref{tab:human_eval} affirm that our MegaFusion significantly improves the performance of higher-resolution image generation in terms of image quality and semantic accuracy.
Additionally, our advanced MegaFusion++ further demonstrates potential for even greater improvements.
This evidence underscores MegaFusion's ability to elevate pre-trained models, enabling them to produce higher-resolution images with superior quality and precise semantics.

\begin{table}[t]
        \small
        \centering
 \setlength
 \tabcolsep{3pt}
 \begin{tabular}{c|c|c|c}
    \toprule
    Methods &  Image Quality & Semantics & Preference  \\ 
    \midrule
    SDM  & 2.60 & 2.05 & 5.42\%  \\
    SDM-MegaFusion   & 3.25 & 4.40 & 12.92\%  \\
    SDM-MegaFusion++  & \textbf{4.25} & \textbf{4.55} & \textbf{81.66\%}  \\ 
    \midrule
    Floyd-Stage2  & 2.18 & 4.28 & 1.67\%  \\
    Floyd-MegaFusion   & 3.45 & \textbf{4.58} & 21.25\%  \\
    Floyd-MegaFusion++  & \textbf{4.22} & \textbf{4.58} & \textbf{77.08\%}  \\ 
    \bottomrule
\end{tabular}
        \vspace{-0.2cm}
        \caption{\textbf{Human evaluation} with MS-COCO captions and commonly used prompts from the Internet as input.}
        \label{tab:human_eval}
        \vspace{-0.4cm}
\end{table}

\subsection{Qualitative Results}
\label{5.3}
\noindent{\textbf{Comparison on text-to-image foundation models.}}
Figure~\ref{fig:qualitative} showcases visualization results of higher-resolution image generation in both latent and pixel spaces.
These results affirm that MegaFusion can be seamlessly integrated with existing diffusion models to produce images of megapixels with accurate semantics, whereas prior baselines fail to do so.
Moreover, incorporating dilated convolutions and noise re-scheduling further improves image details.
Additional results are available in the Appendix.

\vspace{2pt}
\noindent{\textbf{Comparison on models with additional conditions.}}
We further apply MegaFusion to diffusion models equipped with extra input conditions, such as IP-Adapter and ControlNet, as illustrated in Figure~\ref{fig:qualitative2}.
Our MegaFusion exhibits universal applicability, significantly extending the capacity of various diffusion models to synthesize high-quality images of higher resolutions, which not only adhere to the input conditions but also maintain semantic integrity.
Please refer to the Appendix for more qualitative results.
\begin{table}[t]
        \centering
        \small
 \setlength
 \tabcolsep{2pt}
\begin{tabular}{c|ccc|c|c|c|c}
    \toprule
    Methods &  T\&R & D & R & $\mathrm{FID}_r$ & $\mathrm{FID}_b$ & $\mathrm{KID}_r$ & $\mathrm{KID}_b$ \\ 
    \midrule
    SDM~\cite{SDM} & \XSolidBrush & \XSolidBrush & \XSolidBrush & 41.35 & 51.02 & 0.0086 & 0.0113 \\ 
    SDM-MegaFusion & \Checkmark   & \XSolidBrush & \XSolidBrush & 30.19 & 10.98 & 0.0088 & 0.0034 \\ 
    SDM-MegaFusion-D  & \Checkmark & \Checkmark & \XSolidBrush & 27.56 & 9.34 & 0.0075 & 0.0019 \\ 
    SDM-MegaFusion-R  & \Checkmark & \XSolidBrush & \Checkmark & 26.78 & 9.34 & 0.0075 & 0.0019 \\ 
    SDM-MegaFusion++  & \Checkmark & \Checkmark & \Checkmark & \textbf{25.14} & \textbf{7.82} & \textbf{0.0064} & \textbf{0.0012} \\ 
    \midrule
    Floyd-Stage1 & \XSolidBrush & \XSolidBrush & \XSolidBrush & 66.27 & 81.65 & 0.0262 & 0.0454 \\ 
    Floyd-MegaFusion  & \Checkmark & \XSolidBrush & \XSolidBrush & 53.09 & \textbf{39.73} & 0.0273 & \textbf{0.0334} \\ 
    Floyd-MegaFusion-D  & \Checkmark & \Checkmark & \XSolidBrush & 51.76 & 41.96 & 0.0268 & 0.0345 \\ 
    Floyd-MegaFusion-R  & \Checkmark & \XSolidBrush & \Checkmark & 44.27 & 49.38 & 0.0215 & 0.0431 \\ 
    Floyd-MegaFusion++  & \Checkmark & \Checkmark & \Checkmark & \textbf{43.43} & 50.08 & \textbf{0.0213} & 0.0437  \\ 
    \bottomrule
    \end{tabular}
    \vspace{-0.2cm}
    \caption{\textbf{Ablation Study} on proposed modules in MegaFusion on MS-COCO.
    The modules gradually improve the higher-resolution generation quality, especially in comparison with real images.
    }
    \label{tab:strategy_ablation}
    \vspace{-0.5cm} 
\end{table}

\subsection{Ablation Studies}
\label{5.4}
\noindent{\textbf{Proposed strategy \& modules.}}
To evaluate the efficacy of our proposed strategy and components, we assess several model variants in both latent and pixel spaces.
Here, `T\&R', `D', and `R' represent the {\em truncate and relay} strategy, {\em dilated convolution}, and {\em noise re-scheduling}, respectively.
The results in Table~\ref{tab:strategy_ablation} demonstrate that our strategy and modules significantly elevate the quality and diversity of contents generated by generative models such as SDM ($1024 \times 1024$) and Floyd ($128 \times 128$), especially in improving the quality and fidelity to real-world images.

\vspace{2pt}
\noindent{\textbf{Upsampler $\Phi$.}}
The non-parametric Upsampler is crucial in our {\em truncate and relay} strategy to bridge generation processes across different resolutions.
To determine the optimal choice, we evaluate several variants of SDM-MegaFusion++ on MS-COCO dataset, including Config-A~(bilinear upsampling); Config-B (bicubic upsampling); Config-C (bicubic with a $5 \times 5$ Gaussian filter), and Config-D (bicubic with a $3 \times 3$ edge-enhancement kernel).
As depicted in Table~\ref{tab:function_ablation}, SDM-MegaFusion++ with Config-B outperforms others in terms of both FID and KID metrics, leading us to adopt bicubic upsampling as the default choice.

\begin{table}[t]
        \centering
 \setlength
 \tabcolsep{6pt}
\begin{tabular}{c|c|c|c|c}
    \toprule
    Methods &  $\mathrm{FID}_r\downarrow$ & $\mathrm{FID}_b\downarrow$ & $\mathrm{KID}_r\downarrow$ & $\mathrm{KID}_b\downarrow$ \\ 
    \midrule
    SDM  & 41.35 & 51.02 & 0.0086 & 0.0113 \\ 
    \hline
    Config-A     & 28.03 & 9.70 & 0.0076 & 0.0020 \\ 
    Config-B     & \textbf{25.14} & \textbf{7.82} & \textbf{0.0064} & \textbf{0.0012} \\
    Config-C     & 35.07  & 18.10 & 0.0118 & 0.0063 \\
    Config-D  & 26.56 & 13.26 & 0.0065 & 0.0021  \\ 
    \bottomrule
    \end{tabular}
    \vspace{-0.2cm}
    \caption{\textbf{Ablation study} on Upsampler function $\Phi$.}
    \label{tab:function_ablation}
    \vspace{-0.4cm} 
\end{table}

\section{Conclusion}
In this paper, we present \textbf{MegaFusion}, a tuning-free approach designed to tackle the challenges of synthesizing higher-resolution images, effectively resolving issues of semantic inaccuracies and object replication.
Our method adopts an innovative \textbf{{\em truncate and relay}} strategy to elegantly connect generation processes across different resolutions, synthesizing higher-resolution images with megapixels and various aspect ratios.
By integrating {\em dilated convolutions} and {\em noise re-scheduling}, we further improve the synthesis quality.
The versatility of MegaFusion makes it universally applicable to both latent-space and pixel-space diffusion models, as well as their extensions with extra conditions.
Extensive experiments have validated the superiority of MegaFusion, demonstrating its capability to generate higher-resolution images with approximately 40\% of the original computational cost.
\section*{Acknowledgement}
This work is supported by National Natural Science Foundation of China (62271308), STCSM (22511105700, 22DZ2229005), 111 plan (BP0719010), and State Key Laboratory of UHD Video and Audio Production and Presentation.

{\small
\bibliographystyle{ieee_fullname}
\bibliography{egbib}
}

\clearpage
\appendix

In this appendix, we start by giving more details on the implementation details of our proposed MegaFusion in Section~\ref{supp_1}. 
Then, we provide extra quantitative comparisons to further demonstrate the universality and effectiveness of our method in Section~\ref{supp_2}.
Next, we offer additional qualitative results across various experimental settings and methods to illustrate the superiority of our proposed MegaFusion in Section~\ref{supp_3}.
Finally, we discuss the limitations of our method and future work in Section~\ref{supp_4}.

\section{Implementation Details}
\label{supp_1}
\noindent{\textbf{More Details on Floyd-MegaFusion.}}
We have evaluated the higher-resolution image generation performance of Floyd~\cite{DeepFloyd} at resolutions of $128\times 128$ and $512 \times 512$.
For $128 \times 128$ resolution, we directly apply MegaFusion to the first stage of Floyd.
As for the comparison at $512 \times 512$ resolution, we utilize the first two stages of Floyd.
Considering that the quality of the results from the first stage generation would significantly affect the second generation stage, we opt for using the $64 \times 64$ images generated by the original first stage model as inputs of both the baseline and our boosted Floyd-MegaFusion.
That is, higher-resolution image generation is only performed under the second generation stage.
Ultimately, the experimental results presented in Table~\ref{tab:quantitative_results} of our submitted manuscript effectively demonstrate the universality and effectiveness of our proposed MegaFusion.
Furthermore, we also conduct experiments where $128 \times 128$ out-of-distribution images are generated in the first stage, followed by $512 \times 512$ resolution images in the second stage. 
This further demonstrates that MegaFusion maintains semantic accuracy across all stages of generation.

\vspace{2pt}
\noindent{\textbf{Details on Human Evaluation.}}
To more effectively reflect the performance of different models in generating high-resolution images, we have recruited 10 volunteers with a background in image generation research for human evaluation.
Specifically, the evaluators are asked to follow these rules: 
(i) Rate unknown source images on a score from 1 to 5 for both image quality and semantic accuracy, with higher scores indicating better quality;
and (ii) Observe the results generated by different models with the same input conditions and select their favourite one based on overall quality and semantic accuracy.

\section{Additional Quantitative Results}
\label{supp_2}

\subsection{Comparison on crop FID/KID}
Following previous work~\cite{du2024demofusion}, we also evaluate crop FID and crop KID metrics on the generated results of various models to reflect the quality of local patches in the images.
As depicted in Table~\ref{tab:quantitative_results_supp}, previous methods are often limited to specific latent-space models, whereas our MegaFusion consistently improves the quality of high-resolution image generation across both latent-space and pixel-space models.

\begin{table}[tph]
\small
\tabcolsep=0.06cm
\footnotesize
\begin{center}
\begin{tabular}{c|c|c|c|c}
\toprule
Method & SDM-1024 & SDXL-2048 & Floyd-128 & Floyd-512 \\ 
\midrule
Original     & 41.21/0.0139 & 42.29/0.0125 & 70.16/0.0224 & 40.65/0.0171 \\ 
ScaleCrafter   & \textbf{32.24}/0.0085 & 26.58/0.0062 & inapplicable & inapplicable \\ 
DemoFusion   & inapplicable & 25.91/0.0061 & inapplicable & inapplicable \\ 
MegaFusion   & 39.42/0.0137 & 27.38/0.0063 & 57.24/0.0243 & 32.36/0.0122 \\ 
MegaFusion++ & 33.39/\textbf{0.0084} & \textbf{25.64}/\textbf{0.0049} & \textbf{41.22}/\textbf{0.0188} & \textbf{29.18}/\textbf{0.0077} \\ 
\bottomrule
\end{tabular}
\end{center}
\vspace{-0.4cm}
\caption{Comparison of $\mathrm{FID_{crop}}$/$\mathrm{KID_{crop}}$ on MS-COCO dataset.}
\vspace{-0.2cm}
\label{tab:quantitative_results_supp}
\end{table}

\begin{table*}[thp]
\begin{center}
\tabcolsep=0.06cm
\small
\begin{tabular}{c|c|c|c|c|c|c|c|c|c|c|c}
\toprule
Methods & resolution & $\mathrm{FID}_r\downarrow$ & $\mathrm{FID}_b\downarrow$ & $\mathrm{KID}_r\downarrow$ & $\mathrm{KID}_b\downarrow$ & CLIP-T$\uparrow$ & CIDEr$\uparrow$ & Meteor$\uparrow$ & ROUGE$\uparrow$ & GFlops & Inference time \\ 
\midrule
SDM~\cite{SDM} & $1024 \times 1024$ & 77.92 & 46.34 & 0.0363 & 0.0220 & 0.2952 & 8.12 & 7.48 & 7.09 & 135.0K & 15.17s \\ 
SDM-MegaFusion & $1024 \times 1024$ &  \textcolor{blue}{\underline{71.78}} & \textcolor{blue}{\underline{36.21}}  &  \textcolor{blue}{\underline{0.0303}} & \textcolor{blue}{\underline{0.0189}}  & \textcolor{blue}{\underline{0.3060}}  &  \textcolor{blue}{\underline{24.46}} & \textcolor{blue}{\underline{11.98}} &  \textcolor{blue}{\underline{12.62}} & \textbf{48.2K} & \textbf{7.56s} \\
SDM-MegaFusion++ & $1024 \times 1024$ &\textcolor{red}{\textbf{68.92}}  & \textcolor{red}{\textbf{34.94}} &\textcolor{red}{\textbf{0.0251}} &\textcolor{red}{\textbf{0.0182}} & \textcolor{red}{\textbf{0.3115}} &\textcolor{red}{\textbf{28.52}}  & \textcolor{red}{\textbf{12.32}} &\textcolor{red}{\textbf{13.29}} & \textbf{48.2K} & \textbf{7.56s} \\
\midrule

SDXL~\cite{podell2023sdxl} & $2048 \times 2048$ & 73.49 & 48.78 & 0.0308 & 0.0274 & 0.2994 & 16.43 & 9.90 & 10.35 & 540.2K & 79.66s \\ 
SDXL-MegaFusion & $2048 \times 2048$ & \textcolor{blue}{\underline{72.62}} & \textcolor{blue}{\underline{13.72 }} & \textcolor{blue}{\underline{0.0296}} & \textcolor{blue}{\underline{0.0039}} &\textcolor{blue}{\underline{0.3113}}  & \textcolor{blue}{\underline{25.98}} &\textcolor{blue}{\underline{13.23}}  & \textcolor{blue}{\underline{13.33}} & \textbf{216.1K} & \textbf{30.94s} \\ 
SDXL-MegaFusion++ & $2048 \times 2048$ & \textcolor{red}{\textbf{65.10}} &\textcolor{red}{\textbf{11.55}} & \textcolor{red}{\textbf{0.0225}} &\textcolor{red}{\textbf{0.0026}}& \textcolor{red}{\textbf{0.3122}}&\textcolor{red}{\textbf{26.35}} &\textcolor{red}{\textbf{13.98}} &\textcolor{red}{\textbf{14.92}}& \textbf{216.1K} & \textbf{30.94s} \\ 
\midrule
Floyd-Stage1~\cite{DeepFloyd} & $128 \times 128$ & 87.04 & 105.59 &\textcolor{blue}{\underline{0.0341}}  & 0.0658 & 0.2866 & 9.95 & 8.28 & 9.07 & 111.7K & 77.08s \\ 
Floyd-MegaFusion & $128 \times 128$ & \textcolor{blue}{\underline{77.82}}  & \textcolor{red}{\textbf{36.49}} &0.0413 &\textcolor{red}{\textbf{0.0281}} & \textcolor{blue}{\underline{0.3080}}  &\textcolor{blue}{\underline{22.12}}   &\textcolor{red}{\textbf{17.06}}  & \textcolor{red}{\textbf{20.62}} & \textbf{44.9K} & \textbf{32.19s} \\
Floyd-MegaFusion++ & $128 \times 128$ & \textcolor{red}{\textbf{73.54}} &\textcolor{blue}{\underline{45.76}}   &\textcolor{red}{\textbf{0.0334}}  &\textcolor{blue}{\underline{0.0388}}   &\textcolor{red}{\textbf{0.3086}}  &\textcolor{red}{\textbf{22.52}}  &\textcolor{blue}{\underline{16.93}}   &\textcolor{blue}{\underline{20.05}}   & \textbf{44.9K} & \textbf{32.19s} \\
\midrule
Floyd-Stage2~\cite{DeepFloyd} & $512 \times 512$ & 80.34 & 41.65 & 0.0401 & 0.0215 & 0.3013 & 23.59 & 12.28 & 11.67 & 60.7K & 48.58s  \\ 
Floyd-MegaFusion & $512 \times 512$ &\textcolor{blue}{\underline{77.66}}  & \textcolor{blue}{\underline{39.34}} & \textcolor{blue}{\underline{0.0348}} &\textcolor{blue}{\underline{0.0141}}  & \textcolor{blue}{\underline{0.3110}} & \textcolor{blue}{\underline{24.63}} &  \textcolor{red}{\textbf{15.74}} &\textcolor{blue}{\underline{15.29}}  & \textbf{24.3K} & \textbf{21.72s} \\ 
Floyd-MegaFusion++ & $512 \times 512$ &  \textcolor{red}{\textbf{62.91}} &  \textcolor{red}{\textbf{34.40}} &  \textcolor{red}{\textbf{0.0232}} &  \textcolor{red}{\textbf{0.0115}} & \textcolor{red}{\textbf{0.3141}} & \textcolor{red}{\textbf{25.44}} & \textcolor{blue}{\underline{13.90}} & \textcolor{red}{\textbf{18.51}} & \textbf{24.3K} & \textbf{21.72s} \\ 
\bottomrule
\end{tabular}
\end{center}
\vspace{-0.4cm}
\caption{\textbf{Quantitative comparison} on CUB-200~\cite{Wah2011TheCB} dataset. 
\textcolor{red}{\textbf{RED}}: best performance, \textcolor{blue}{\underline{BLUE}}: second best performance.
}
 \label{tab:quantitative_results_2}
\end{table*}

\begin{table*}[thp]
\begin{center}
\tabcolsep=0.08cm
\small
\begin{tabular}{c|c|c|c|c|c|c|c|c|c|c|c}
\toprule
Methods & resolution & $\mathrm{FID}_r\downarrow$ & $\mathrm{FID}_b\downarrow$ & $\mathrm{KID}_r\downarrow$ & $\mathrm{KID}_b\downarrow$ & CLIP-T$\uparrow$ & CIDEr$\uparrow$ & Meteor$\uparrow$ & ROUGE$\uparrow$ & GFlops & Inference time \\ 
\midrule
Floyd-Stage1~\cite{DeepFloyd} & $128 \times 128$ & 66.27 & 81.65 & \textcolor{blue}{\underline{0.0262}} & 0.0454 & 0.2818 & 14.69 & 18.22 & 25.06 & 111.7K & 77.08s \\ 
Floyd-MegaFusion & $128 \times 128$ & \textcolor{blue}{\underline{53.09}} & \textcolor{red}{\textbf{39.73}} & 0.0273 & \textcolor{red}{\textbf{0.0334}} & \textcolor{blue}{\underline{0.3024}} & \textcolor{red}{\textbf{25.01}} & \textcolor{blue}{\underline{25.00}} & \textcolor{blue}{\underline{31.35}} & \textbf{44.9K} & \textbf{32.19s} \\
Floyd-MegaFusion++ & $128 \times 128$ & \textcolor{red}{\textbf{43.43}} & \textcolor{blue}{\underline{50.08}} & \textcolor{red}{\textbf{0.0213}} & \textcolor{blue}{\underline{0.0437}} & \textcolor{red}{\textbf{0.3046}} & \textcolor{blue}{\underline{20.28}} & \textcolor{red}{\textbf{25.01}} & \textcolor{red}{\textbf{31.64}} & \textbf{44.9K} & \textbf{32.19s} \\
\midrule
Floyd-Stage2~\cite{DeepFloyd} & $64 \rightarrow 512$ & 46.64 & 38.15 & 0.0254 & 0.0166 & 0.3098 & \textcolor{blue}{\underline{23.85}} & 21.47 & 26.26 & 60.7K & 48.58s  \\ 
Floyd-MegaFusion & $64 \rightarrow 512$ & \textcolor{blue}{\underline{39.80}} & \textcolor{blue}{\underline{24.87}} & \textcolor{blue}{\underline{0.0164}} &\textcolor{blue}{\underline{0.0078}} & \textcolor{blue}{\underline{0.3106}} & 23.22 & \textcolor{blue}{\underline{23.51}} & \textcolor{blue}{\underline{29.30}} & \textbf{24.3K} & \textbf{21.72s} \\ 
Floyd-MegaFusion++ & $64 \rightarrow 512$ & \textcolor{red}{\textbf{26.34}} & \textcolor{red}{\textbf{24.55}} & \textcolor{red}{\textbf{0.0063}} & \textcolor{red}{\textbf{0.0077}} & \textcolor{red}{\textbf{0.3110}} & \textcolor{red}{\textbf{24.01}} & \textcolor{red}{\textbf{23.58}} & \textcolor{red}{\textbf{29.52}} & \textbf{24.3K} & \textbf{21.72s} \\ 
\midrule
Floyd-Stage2~\cite{DeepFloyd} & $128 \rightarrow 512$ & 61.24 & 108.01 & 0.0253 & 0.0734 & 0.2779 & 15.16 & 14.76 & 19.75 & 60.7K & 48.58s  \\ 
Floyd-MegaFusion & $128 \rightarrow 512$ &\textcolor{blue}{\underline{58.19}}  & \textcolor{red}{\textbf{88.56}} & \textcolor{blue}{\underline{0.0187}} &\textcolor{red}{\textbf{0.0379}}  & \textcolor{blue}{\underline{0.2821}} & \textcolor{blue}{\underline{16.28}} &  \textcolor{red}{\textbf{15.65}} &\textcolor{blue}{\underline{20.02}}  & \textbf{24.3K} & \textbf{21.72s} \\ 
Floyd-MegaFusion++ & $128 \rightarrow 512$ &  \textcolor{red}{\textbf{57.92}} &  \textcolor{blue}{\underline{94.93}} &  \textcolor{red}{\textbf{0.0181}} &  \textcolor{blue}{\underline{0.0417}} & \textcolor{red}{\textbf{0.2835}} & \textcolor{red}{\textbf{16.36}} & \textcolor{blue}{\underline{15.47}} & \textcolor{red}{\textbf{21.34}} & \textbf{24.3K} & \textbf{21.72s} \\ 
\bottomrule
\end{tabular}
\end{center}
\vspace{-0.4cm}
\caption{\textbf{More comparison results} on Floyd model and its MegaFusion boosted counterparts under different settings. 
Within each unit, we denote the best performance in \textcolor{red}{\textbf{RED}} and the second-best performance in \textcolor{blue}{\underline{$\mathrm{BLUE}$}}.
}
\label{tab:quantitative_results_3}
\end{table*}

\subsection{Comparison on CUB-200 Dataset}
To demonstrate the universality of our proposed MegaFusion, in addition to the MS-COCO~\cite{lin2014microsoft} dataset, we also conduct quantitative evaluations on the CUB-200~\cite{Wah2011TheCB} dataset, which is also commonly used in previous works.
The CUB-200 dataset consists of over 10K images of 200 categories of birds, each accompanied by 10 textual descriptions.
Considering computational costs and time expenditure, similar to the experimental settings on the MS-COCO dataset in our manuscript, we randomly select 1K images from the CUB-200 dataset.
Each image is assigned a fixed caption, and the same random seed is used across different methods to eliminate the effects of randomness among models.
As depicted in Table~\ref{tab:quantitative_results_2}, our proposed MegaFusion can also be universally applied to both latent-space and pixel-space diffusion models on the CUB-200 dataset, achieving high-quality higher-resolution image generation.

\subsection{More Results of Floyd-MegaFusion}
As mentioned above, we also conduct experiments that first generate $128 \times 128$ out-of-distribution images, followed by $512 \times 512$ high-resolution images on the Floyd model.
As depicted in Table~\ref{tab:quantitative_results_3}, MegaFusion consistently improves the high-resolution generation capability of Floyd under both settings.
This demonstrates that MegaFusion can improve the semantic accuracy of high-resolution images at any stage of the generation process.

\subsection{Ablation Study of Classifier-free Guidance}
As detailed in the implementation details, to ensure a fair comparison and eliminate the impact of classifier-free guidance~(CFG) on generation quality and efficiency, we use the default CFG weights from official implementations for all methods and their corresponding MegaFusion-boosted counterparts. 
To further investigate the impact of CFG on MegaFusion at higher resolutions, we generate 100 images from the MS-COCO dataset using SDM-MegaFusion and SDXL-MegaFusion with varying CFG values, using the same text prompt and random seed as inputs, and evaluate the FID scores against our testset.
The results in Figure~\ref{fig:CFG} indicate that classifier-free guidance does affect our high-resolution generation quality, with preliminary findings indicating that $w = 7.0$ is a relatively good choice for SDM-MegaFusion and SDXL-MegaFusion.

\begin{figure}[h]
  \centering
  \includegraphics[width=.48\textwidth]{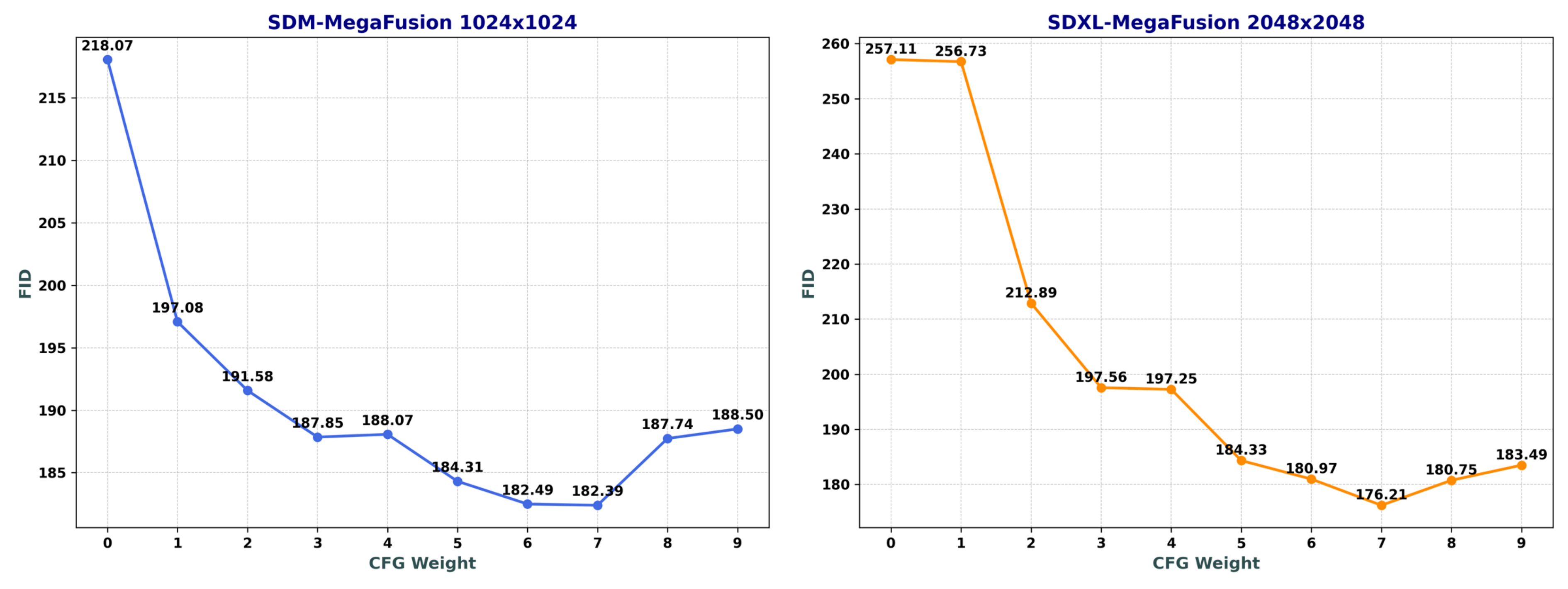} \\
  \vspace{-0.2cm}
  \caption{
  \textbf{Ablation study} of classifier-free guidance (CFG) weight on SDM-MegaFusion and SDXL-MegaFusion.
  }
    \vspace{-0.2cm}
 \label{fig:CFG}
\end{figure}

\section{Additional Qualitative Results}
\label{supp_3}

\subsection{Evidence Behind the Core idea \& intuition}

As stated in eDiff-I~\cite{balaji2022eDiff-I}, diffusion models synthesize semantics during early denoising stages and refine image details in later stages.
As depicted in Figure~\ref{fig:intuition}, we also observe that semantic deviations and object repetitions commonly encountered at higher resolutions primarily stem from incorrect semantics generated during early denoising, leading to irreparable errors.
Thus, our \textbf{intuition and insight} here are: perform early denoising at the original resolution to generate accurate semantic information, followed by {\em truncate} and {\em relay} to continue denoising at higher resolutions, thereby enriching texture details. 
This enables MegaFusion to produce high-quality, semantically accurate higher-resolution images with lower computational costs, while supporting arbitrary aspect ratios.

\begin{figure}[h]
  \centering
  \includegraphics[width=.46\textwidth]{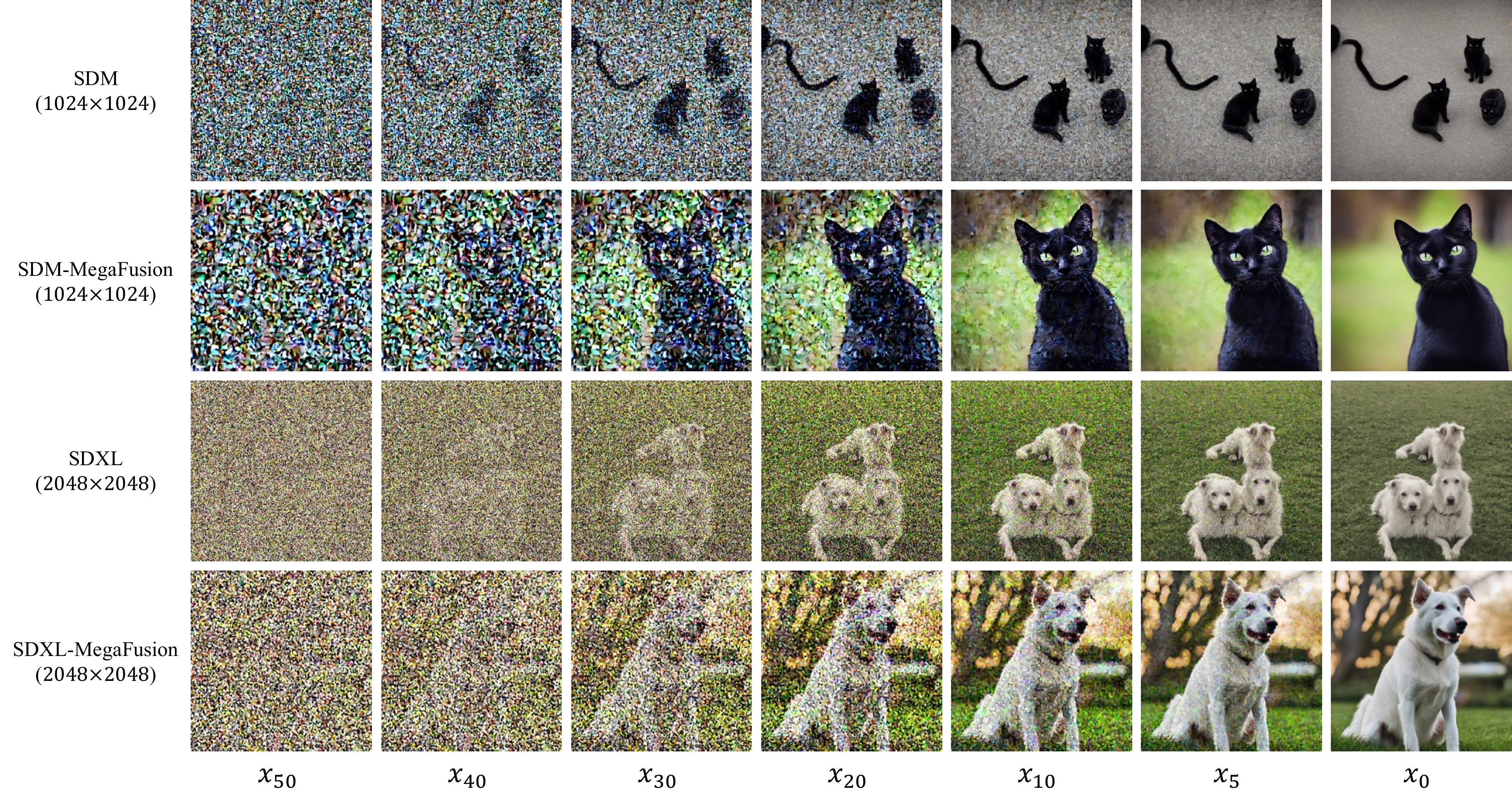} \\
  \vspace{-0.2cm}
  \caption{
  \textbf{Evidence behind our core idea and intuition}.
  For $T=50$ steps of DDIM sampling, we visualize the key stages of the image generation process. 
  For SDM and SDXL, incorrect semantics are generated during the early denoising stages of high-resolution generation, leading to irreparable errors. 
  In contrast, MegaFusion generates accurate semantics and further enriches texture details at higher resolutions. 
  The input text prompts are {``\em A cute black cat''} and {\em ``A white dog sits on the grass.''} 
  For ease of visualization, the images are scaled to the same size.
  }
    \vspace{-0.2cm}
 \label{fig:intuition}
\end{figure}

\subsection{Disadvantages of Direct Upsampling}

Compared to our MegaFusion for higher-resolution image generation, a more straightforward approach is to directly apply upsampling to images generated by diffusion models.
Although simple, this will introduce three potential issues:
(i) Direct super-resolution may lead to unrealistic texture details, such as blurring and artifacts, especially at high upsampling factors;
(ii) While diffusion-based SR methods can produce more realistic textures via iterative denoising, they often involve significantly higher computational costs and may not support arbitrary aspect ratios;
(iii) Most critically, as shown in Figure~\ref{fig:upsampling},  directly upsampling~\cite{wang2021realesrgan, wang2023stablesr, LAR-SR, wang2018esrgan} low-resolution images can stretch and distort content, particularly when generating under non-standard aspect ratios (e.g. $1:4$), diminishing the natural aesthetic of images.

In contrast, MegaFusion seamlessly bridges coarse-to-fine generation processes, efficiently producing accurate semantics at low resolutions and enriching texture details at high resolutions.
Leveraging iterative denoising at higher resolutions, it can synthesize aesthetically pleasing high-resolution images even with non-standard aspect ratios.

\begin{figure}[h]
  \centering
  \includegraphics[width=.48\textwidth]{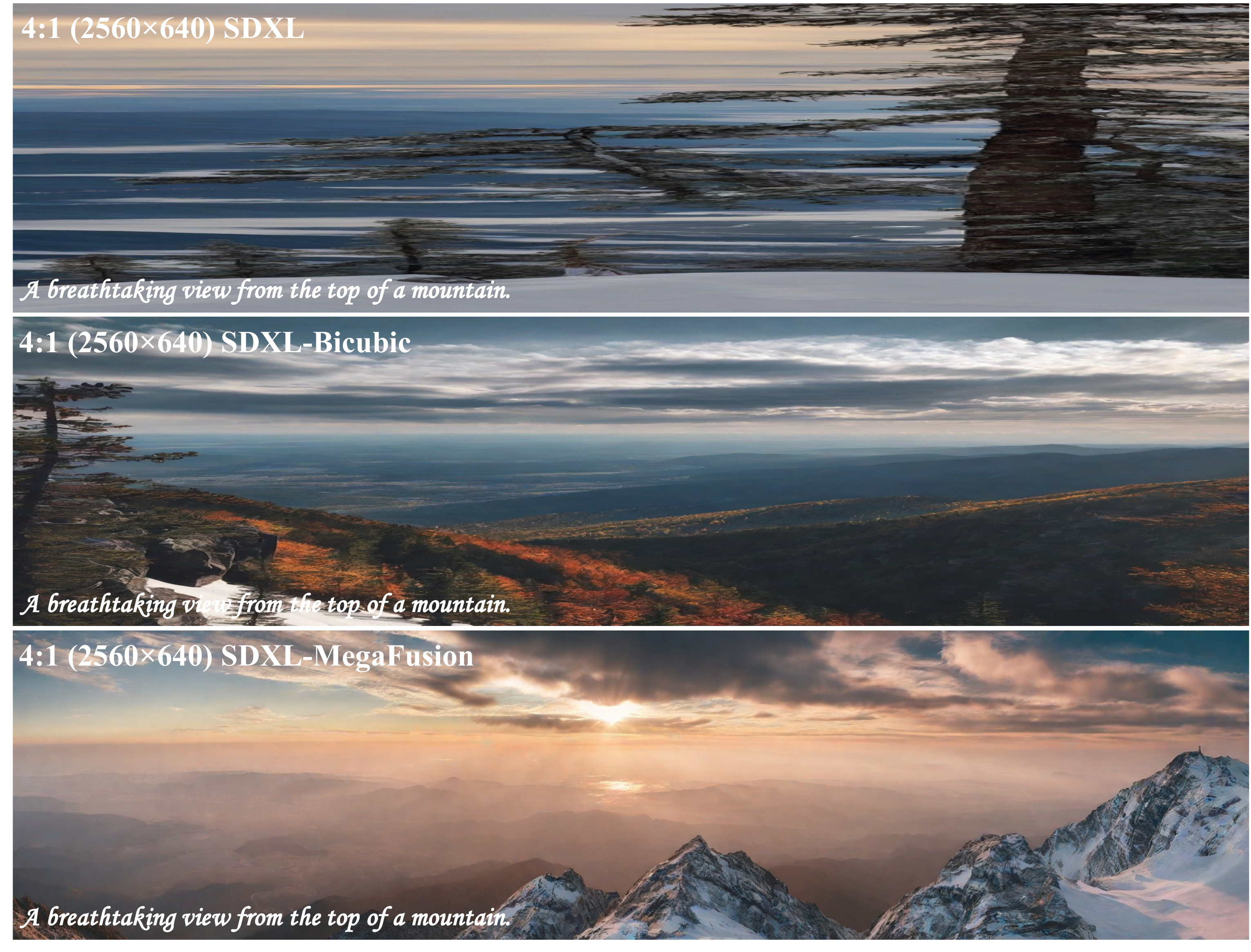} \\
  \vspace{-0.2cm}
  \caption{
  \textbf{Analysis of direct upsampling}. 
  Using diffusion models to generate images with non-standard aspect ratios directly or via upsampling, may lead to stretching and distortion (e.g., trees on both sides), while MegaFusion effectively mitigates this issue.
  }
    \vspace{-0.2cm}
 \label{fig:upsampling}
\end{figure}

\subsection{Effects of hyperparameters $\delta$ and $\gamma$}
For denoising at the original size, we do not employ dilation.
In qualitative experiments for high-resolution generation, we test various $\delta$ values and find that $\delta=2$ is a stable choice under our experimental settings, which will not introduce blurriness or semantic deviations.
As described in our manuscript, we draw inspiration from simple diffusion~\cite{hoogeboom2023simple}, which derives the SNR relationship between images of different resolution based on the mean and variance of pixel distributions.
Substituting this into our derived relationship, we obtain that $\gamma = 4$.
Qualitative experiments also confirm that this is an appropriate choice.
Some visualization examples are shown in Figure~\ref{fig:hyperparameter}.

\begin{figure}[h]
  \centering
  \includegraphics[width=.48\textwidth]{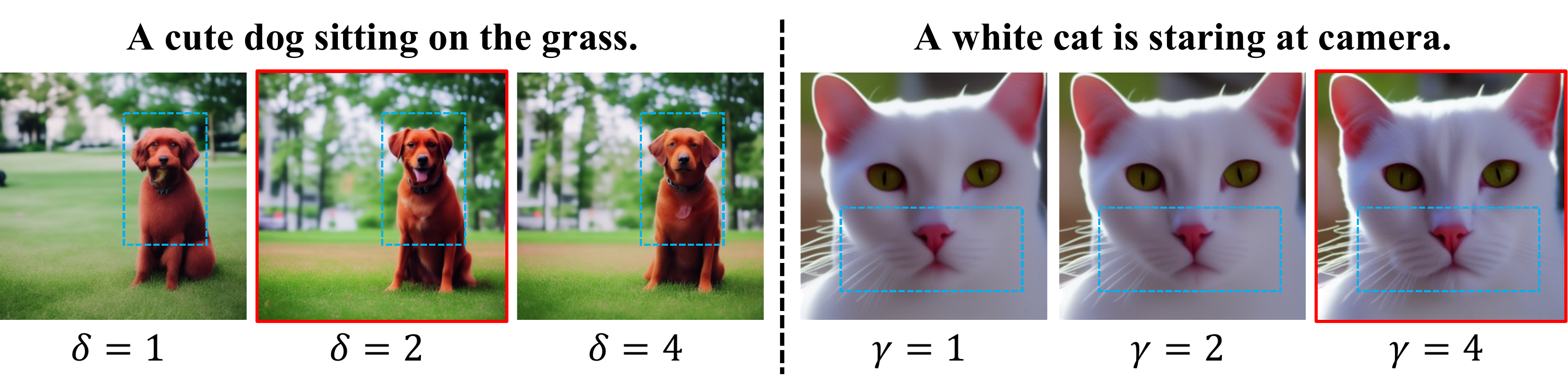} \\
  \vspace{-0.2cm}
  \caption{
  \textbf{Qualitative comparisons} of applying different hyperparameters $\delta$ and $\gamma$.
  }
    \vspace{-0.4cm}
 \label{fig:hyperparameter}
\end{figure}

\subsection{Ablation Study of Truncation Steps}

In the {\em truncate and relay} strategy, the number of denoising steps at each stage may also affect generation quality. 
Our intuition and experience suggest that more denoising steps at lower resolutions improve generation efficiency, while additional steps at higher resolutions enhance texture details. 
However, conducting a comprehensive evaluation to determine the optimal truncation steps would incur significant computational costs. 
Therefore, in our implementation, we empirically select truncation steps for each model based on experience, and validate the above conclusions through qualitative experiments, as shown in Figure~\ref{fig:truncation}.
Considering the trade-off between generation quality and efficiency, we choose denoising steps of $T_1 = 40$, $T_2 = 5$, and $T_3 = 5$ as the default configuration for SDM-MegaFusion.

\begin{figure}[h]
  \centering
  \includegraphics[width=.48\textwidth]{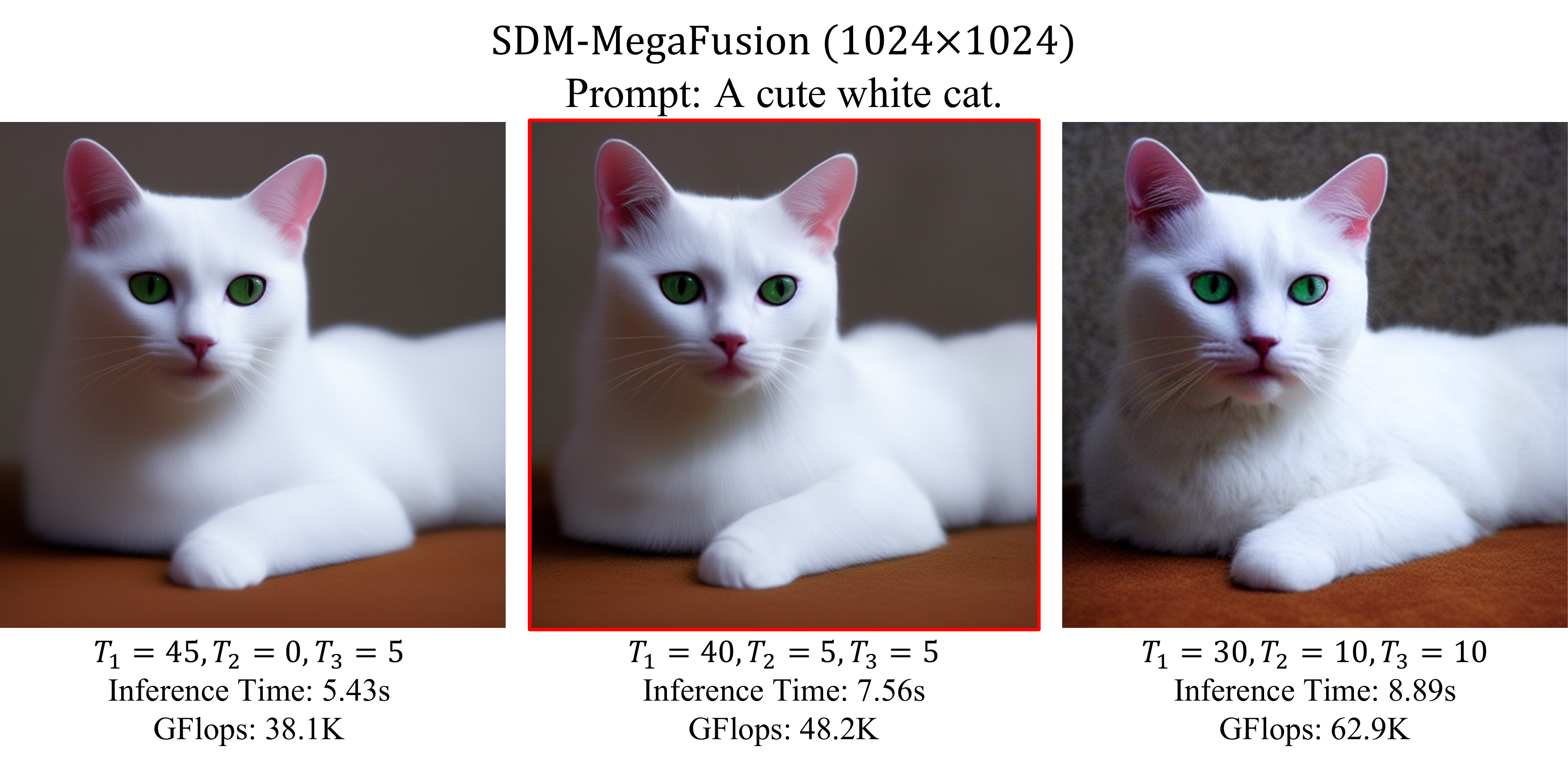} \\
  \vspace{-0.2cm}
  \caption{
  \textbf{Qualitative ablation study} of truncation steps.}
    \vspace{-0.2cm}
 \label{fig:truncation}
\end{figure}

\subsection{Text-to-Image Foundation Models}
We present more visualizations of higher-resolution image generation using both latent-space and pixel-space text-to-image models in Figure~\ref{fig:SDM} and \ref{fig:Floyd}, respectively, to demonstrate the universality and robustness of our proposed method.
The visual outcomes explicitly confirm that when pre-trained models fail to scale to higher resolutions, our approach can be universally integrated into existing latent-space and pixel-space diffusion models, improving their capability to synthesize higher-resolution images of megapixels with accurate semantics.
Moreover, our further enhanced MegaFusion++ significantly boosts the quality of the generated images, producing sharper and clearer details.

\subsection{Compatibility with Transformer-based Models}
To further demonstrate the versatility and effectiveness of MegaFusion, we also apply it to the transformer-based (DiT) SD3~\cite{esserSD3} model. 
Since DiT-based methods do not involve convolutions, we boost the model via only the {\em truncate and relay} strategy.
As shown in Figure~\ref{fig:SD3}, SD3 also encounters quality degradation when generating higher-resolution images directly, while our MegaFusion effectively improves its high-resolution generation capabilities.

\subsection{Comparison to state-of-the-art}
To further evaluate the quality of MegaFusion, we compare it with existing state-of-the-art high-resolution generation methods. 
Given that these methods (ScaleCrafter~\cite{he2023scalecrafter} and DemoFusion~\cite{du2024demofusion}) are typically restricted to specific models, we conduct comparisons on models based on SDXL. 
The results in Figure~\ref{fig:state_of_the_art} indicate that existing methods still face quality degradation and object repetition when generating high-resolution images. 
In contrast, MegaFusion produces high-quality, semantically accurate high-resolution images, and is much more efficient than existing approaches, as shown in Table~\ref{tab:quantitative_results} of our manuscript.

\subsection{Models with additional conditions}
We have confirmed that our method is equally applicable to diffusion models with additional input conditions, such as ControlNet~\cite{controlnet} with depth maps and IP-Adapter~\cite{ye2023ip-adapter} with reference images as extra inputs.
As depicted in Figure~\ref{fig:Controlnet}, we further discover that ControlNet with canny edges or human poses as conditional inputs also struggle with synthesizing higher-resolution images, and often produce images that are not fidelity to input conditions, with confusing semantics and poor image quality.
In contrast, with the assistance of our proposed MegaFusion, our boosted model, ControlNet-MegaFusion consistently generates high-quality images of higher resolutions with accurate semantics, that are fidelity to conditions.

\subsection{Generation with Arbitrary Aspect Ratios}
As previously stated, our MegaFusion also enables existing pre-trained diffusion models to generate images at arbitrary aspect ratios.
Figure~\ref{fig:SDXL_1}, \ref{fig:SDXL_2} and \ref{fig:SDXL_3} showcase more qualitative results from SDXL-MegaFusion across various aspect ratios and resolutions, including $1:1$~($2048\times 2048$), $16:9$~($1920\times 1080$), $3:4$~($1536\times 2048$), and $4:3$ ($2048\times 1536$).
Moreover, as presented in Figure~\ref{fig:SDXL_4}, \ref{fig:SDXL_5}, and \ref{fig:SDXL_6}, we also include visualizations with \textbf{non-standard} aspect ratios, such as $1:4$~($640\times 2560$), $4:1$~($2560\times 640$), $1:2$~($1024\times 2048$), $2:1$~($2048\times 1024$), $21:9$~($2016 \times 864$), and $9:21$ ($864\times 2016$).
These impressive outcomes further demonstrate the scalability and superiority of our approach.

\subsection{Compatibility with LoRA}
To further illustrate the versatility and broad applicability of MegaFusion, we apply it to SDM and SDXL models using LoRA from the open-source community for personalized higher-resolution image generation.
As depicted in Figure~\ref{fig:LoRA}, MegaFusion can seamlessly integrate with various LoRAs of SDM and SDXL, demonstrating significant potential for artistic and commercial applications.

\section{Limitations \& Future Work}
\label{supp_4}
\subsection{Limitations}
Since our proposed MegaFusion is a tuning-free approach built on existing latent-space and pixel-space image generation models, it inevitably inherits some limitations of current diffusion-based generative models.
For example, when handling complex textual conditions, the generated content often struggles to accurately reflect input prompts, particularly in aspects such as attribute binding and positional control.
This may lead to degraded synthesis quality during high-resolution generation with MegaFusion.
However, more powerful backbone models are expected to mitigate this issue, and when combined with MegaFusion, they are likely to produce higher-quality images at higher resolutions with low computational costs.

\subsection{Future Work}
The striking quantitative results produced by MegaFusion have confirmed its potential to overcome the limitations of existing diffusion-based generative models and to improve their capabilities to synthesize high-resolution outcomes.
Additionally, we have observed that existing video generation models encounter significant semantic deviations and quality degradation when generating content beyond their pre-trained spatial resolution and temporal length.
Therefore, we anticipate further applying MegaFusion to current video generation models towards efficient, low-cost, higher-resolution, and longer video content generation.
Similarly, MegaFusion also holds the potential for extension to 3D generation models and models for image and video editing, which are also left for future exploration.

\begin{figure*}[hp]
  \centering
  \includegraphics[width=.9\textwidth]{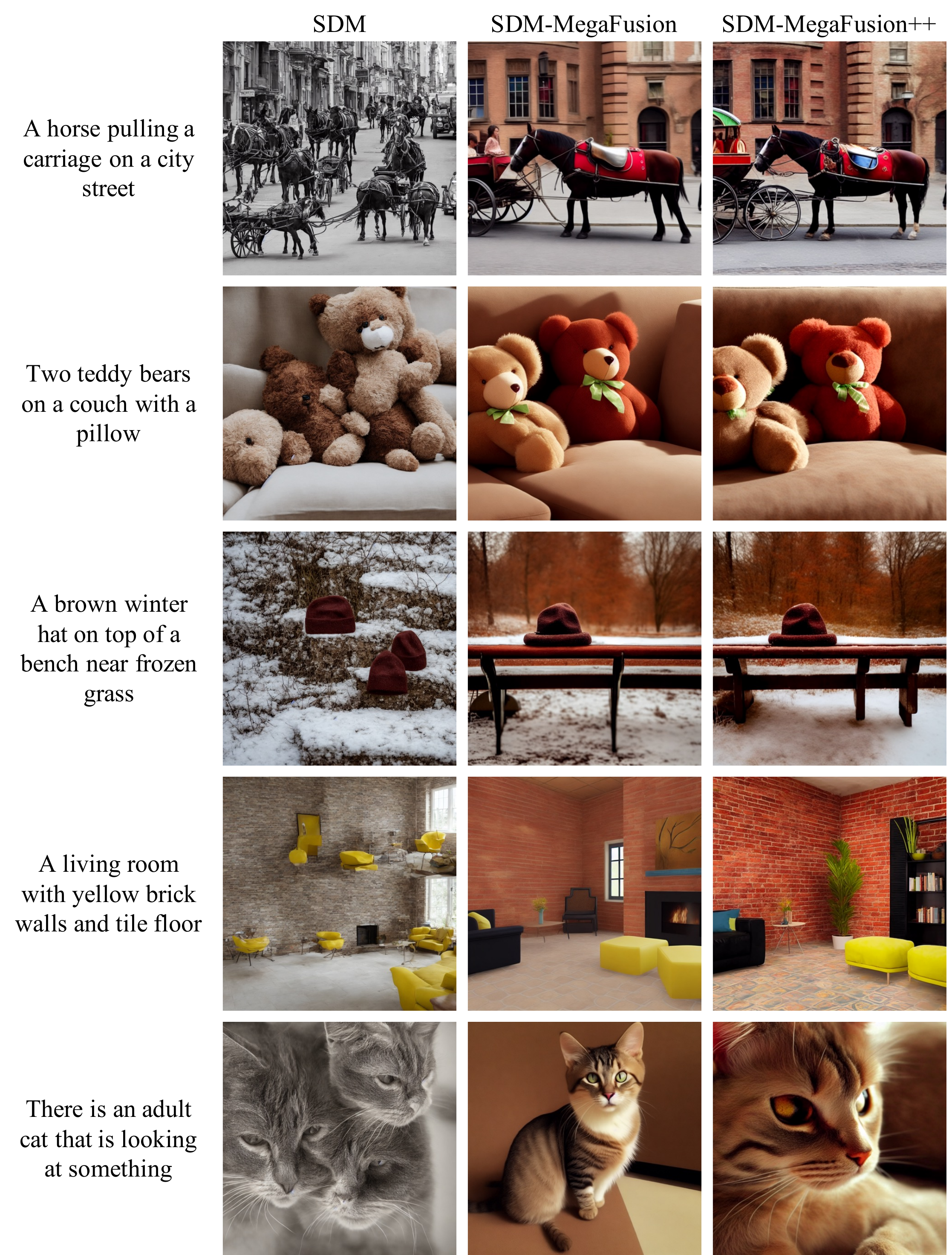} \\
  \caption{
  \textbf{More qualitative results} of applying our MegaFusion to latent-space diffusion model (SDM~\cite{SDM}) for higher-resolution ($1024\times 1024$) image generation on MS-COCO~\cite{lin2014microsoft} and commonly used prompts from the Internet.
  }
 \label{fig:SDM}
\end{figure*}

\begin{figure*}[hp]
  \centering
  \includegraphics[width=.9\textwidth]{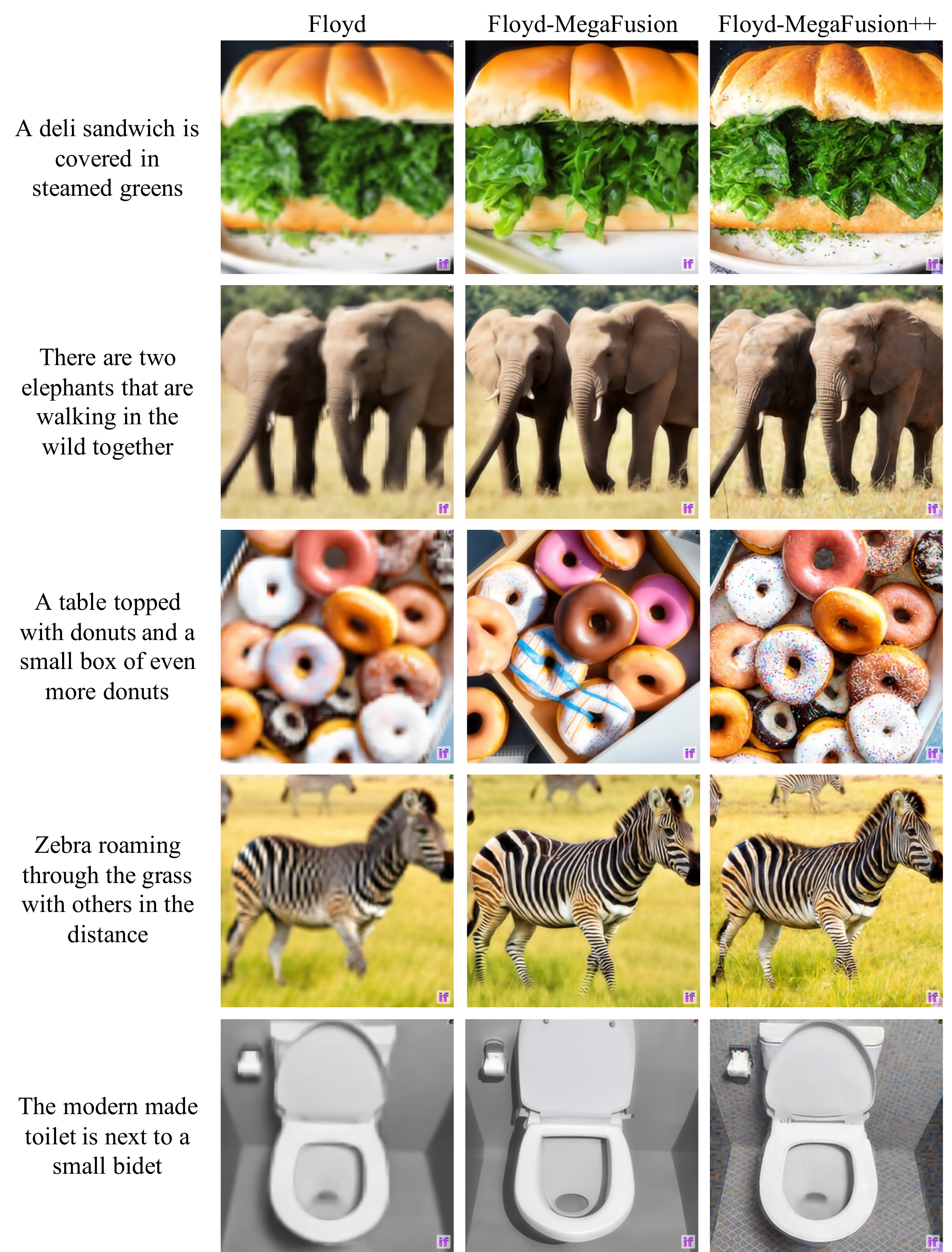} \\
  \caption{
  \textbf{More qualitative results} of applying our MegaFusion to pixel-space diffusion model (Floyd~\cite{DeepFloyd}) for higher-resolution ($512\times 512$) image generation on MS-COCO~\cite{lin2014microsoft} and commonly used prompts from the Internet.
  }
 \label{fig:Floyd}
\end{figure*}

\begin{figure*}[h]
  \centering
  \includegraphics[width=.54\textwidth]{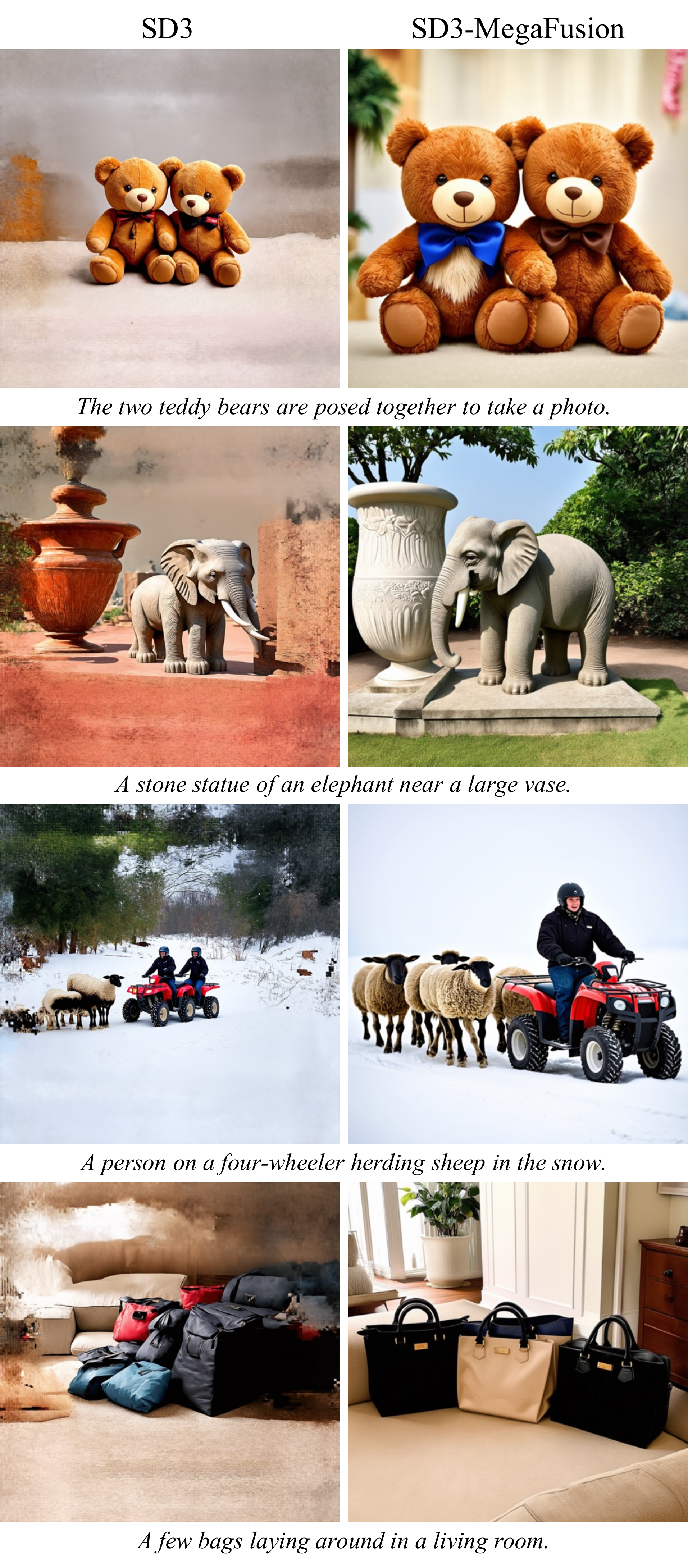} \\
  \vspace{-6pt}
  \caption{
  \textbf{Qualitative results} of applying our MegaFusion to latent-space diffusion model (SD3~\cite{esserSD3}) for higher-resolution ($2048\times 2048$) image generation on MS-COCO~\cite{lin2014microsoft} and commonly used prompts from the Internet.
  }
 \label{fig:SD3}
\end{figure*}

\begin{figure*}[h]
  \centering
  \includegraphics[width=.9\textwidth]{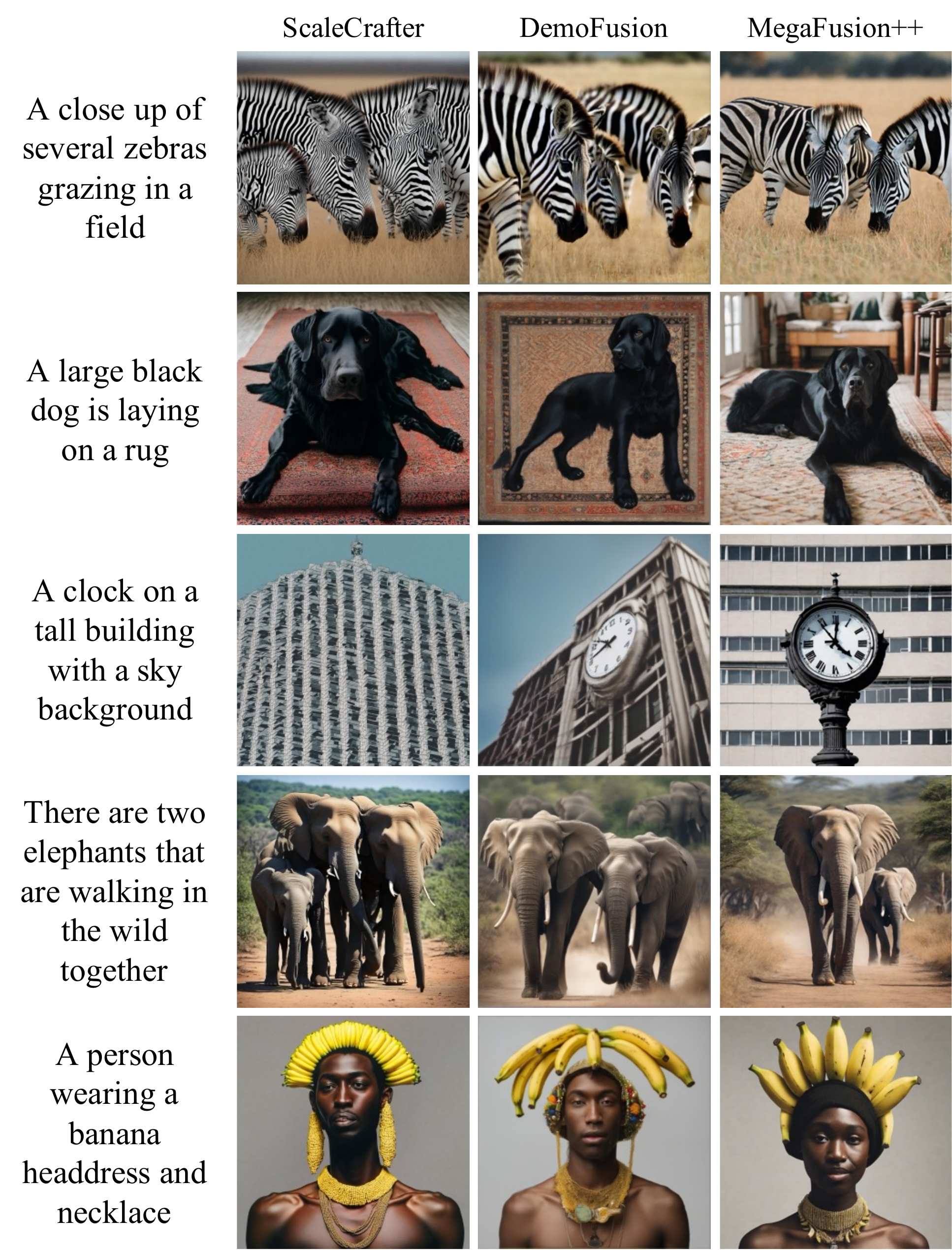} \\
  \caption{
  \textbf{Qualitative comparison} with existing state-of-the-art methods (ScaleCrafter~\cite{he2023scalecrafter} and DemoFusion~\cite{du2024demofusion}).
  Our MegaFusion can generate images with details and accurate semantics at high resolution, whereas existing methods struggle to do so.
  }
 \label{fig:state_of_the_art}
\end{figure*}

\begin{figure*}[h]
  \centering
  \includegraphics[width=.9\textwidth]{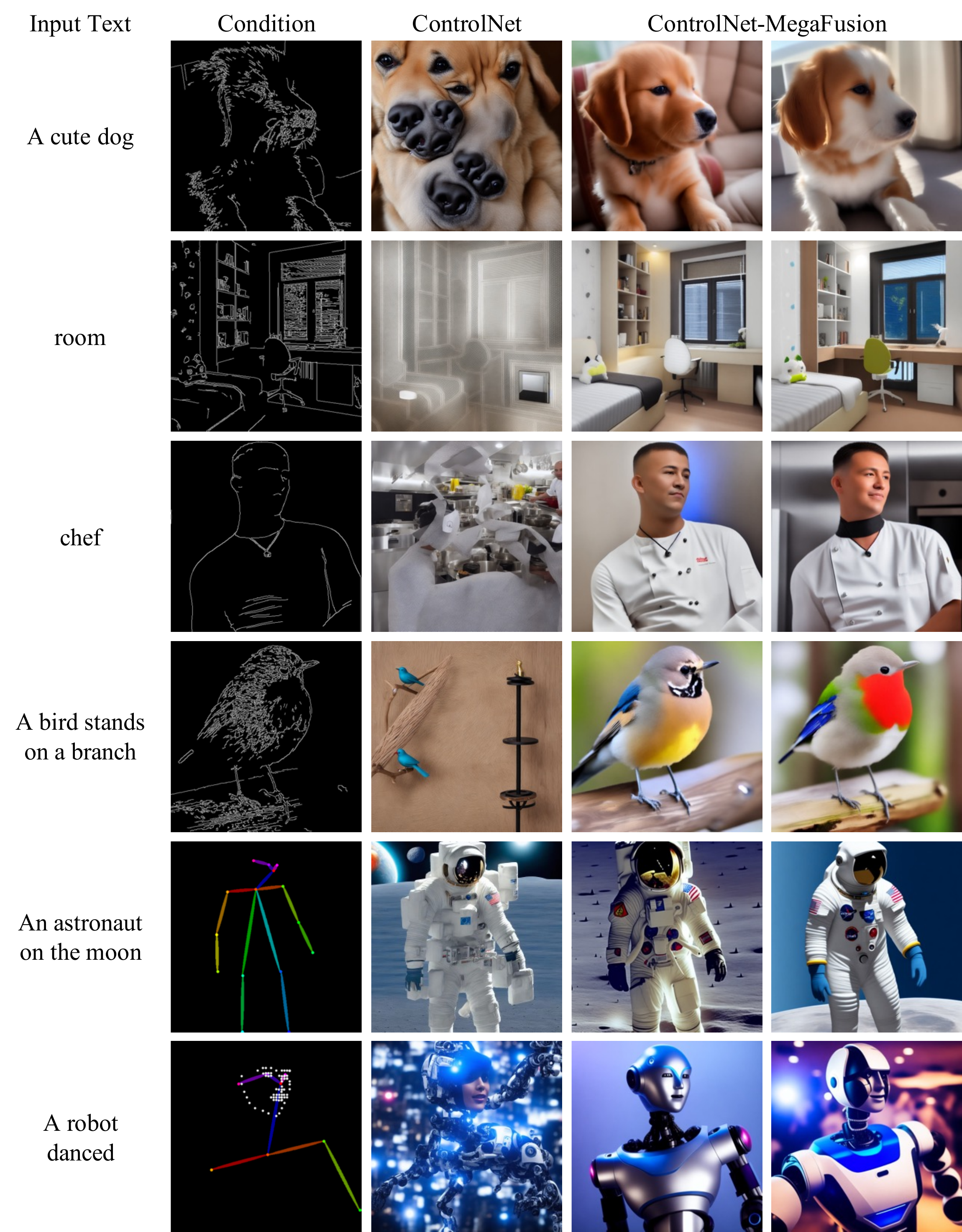} \\
  \caption{
  \textbf{Qualitative results} of applying our MegaFusion to ControlNet~\cite{controlnet} with canny edges or human poses as extra conditions for higher-resolution ($1024\times 1024$) image generation with better semantics and fidelity.
  }
 \label{fig:Controlnet}
\end{figure*}

\begin{figure*}[h]
  \centering
  \includegraphics[width=\textwidth]{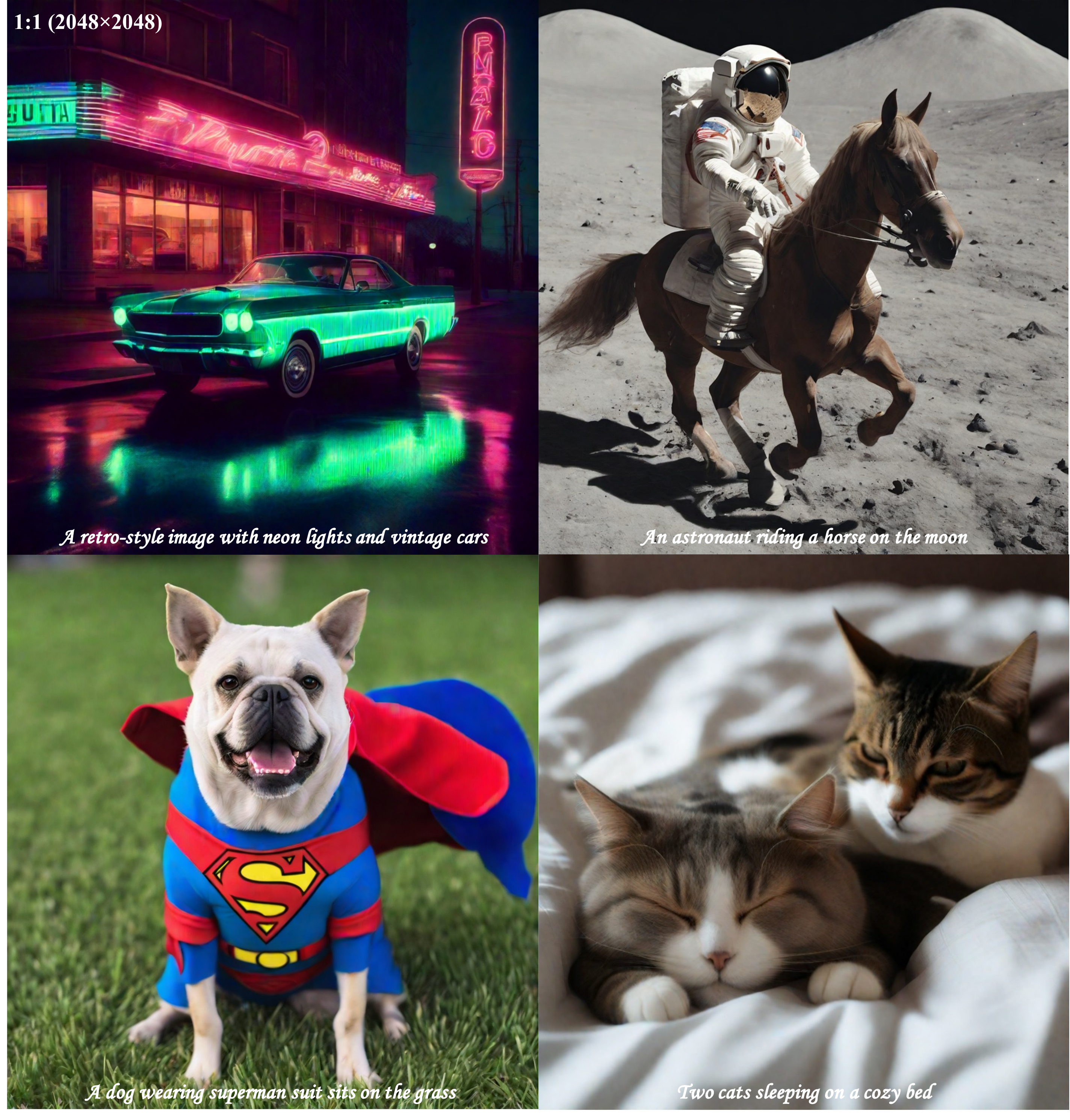} \\
  \caption{
  \textbf{More qualitative results} of applying our MegaFusion to SDXL~\cite{podell2023sdxl} model for higher-resolution image generation with various aspect ratios and resolutions.
  }
 \label{fig:SDXL_1}
\end{figure*}

\begin{figure*}[h]
  \centering
  \includegraphics[width=.7\textwidth]{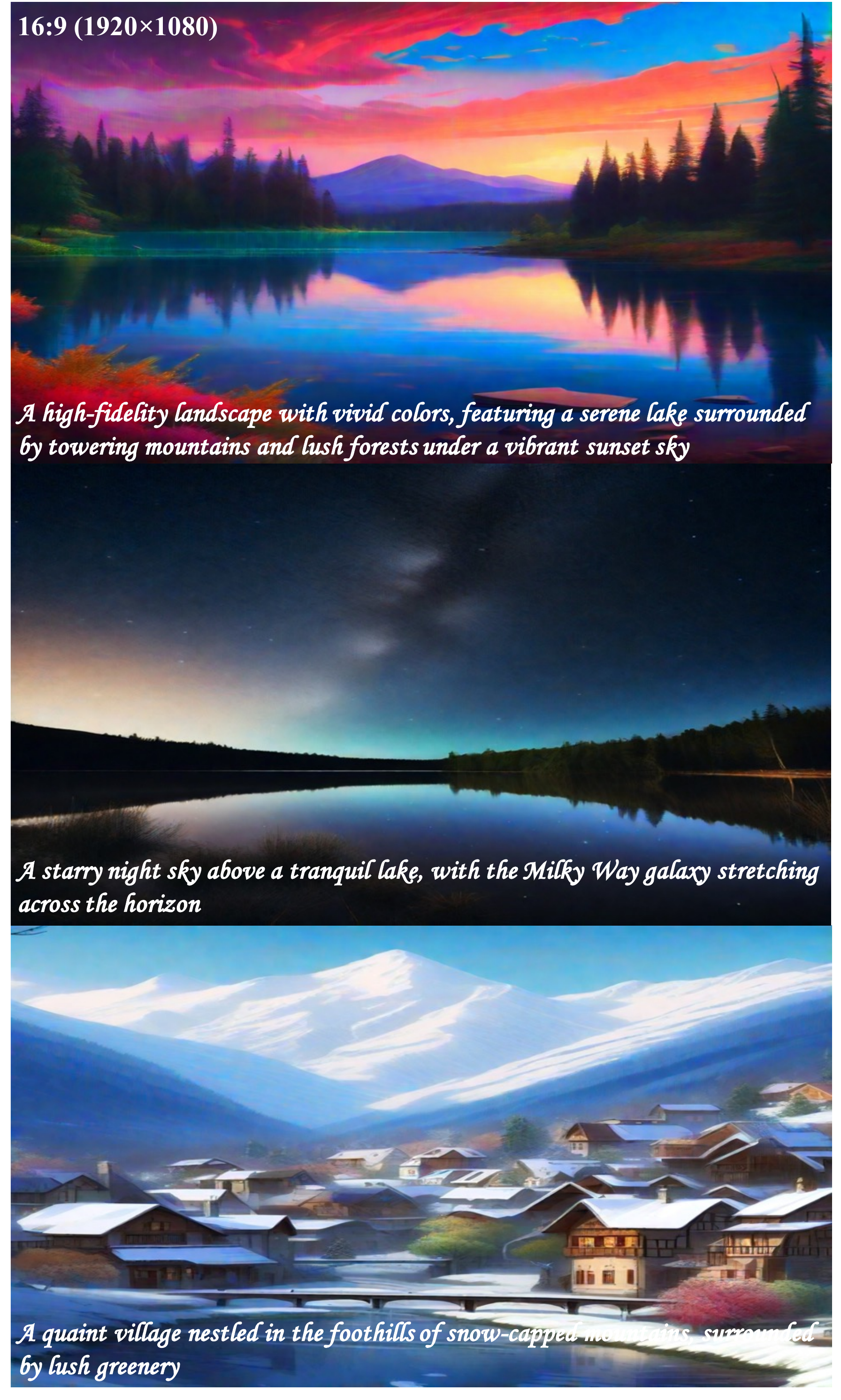} \\
  \caption{
    \textbf{More qualitative results} of applying our MegaFusion to SDXL~\cite{podell2023sdxl} model for higher-resolution image generation with various aspect ratios and resolutions.
  }
 \label{fig:SDXL_2}
\end{figure*}

\begin{figure*}[h]
  \centering
  \includegraphics[width=.7\textwidth]{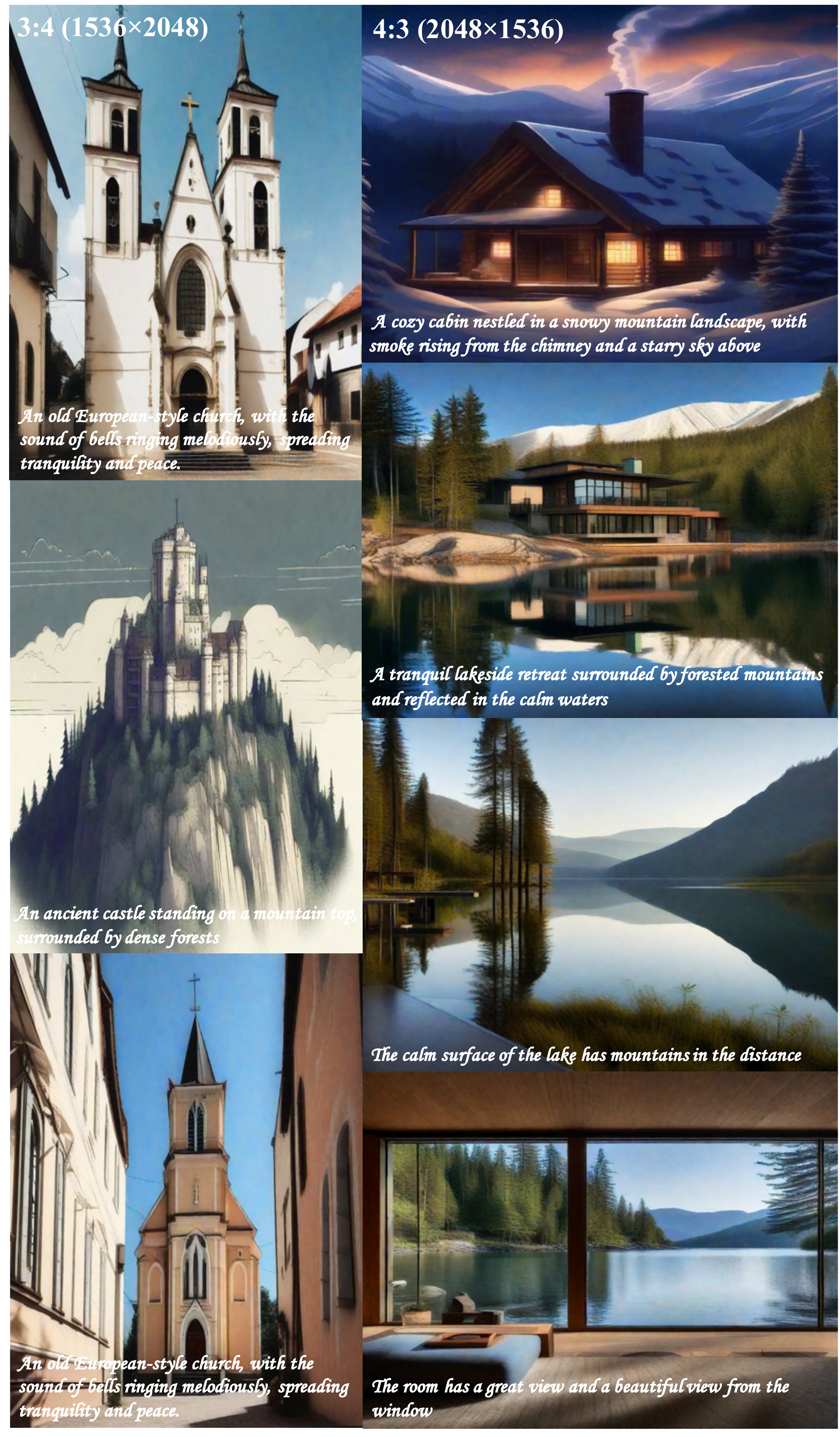} \\
  \caption{
  \textbf{More qualitative results} of applying our MegaFusion to SDXL~\cite{podell2023sdxl} model for higher-resolution image generation with various aspect ratios and resolutions.
  }
 \label{fig:SDXL_3}
\end{figure*}

\begin{figure*}[h]
  \centering
  \includegraphics[width=\textwidth]{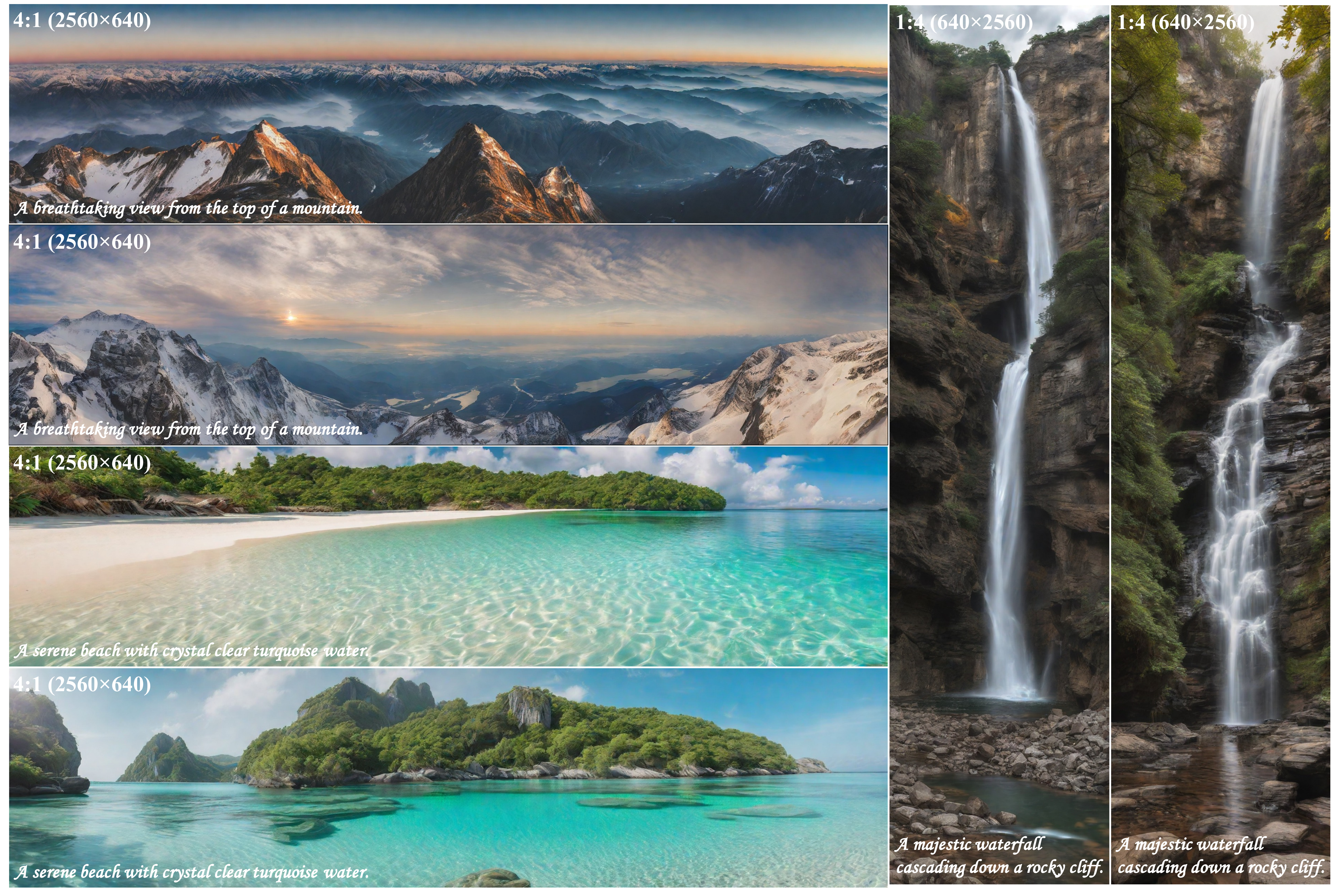} \\
  \caption{
  \textbf{More qualitative results} of applying our MegaFusion to SDXL~\cite{podell2023sdxl} model for higher-resolution image generation with various \textbf{non-standard} aspect ratios and resolutions.}
 \label{fig:SDXL_4}
\end{figure*}

\begin{figure*}[h]
  \centering
  \includegraphics[width=.9\textwidth]{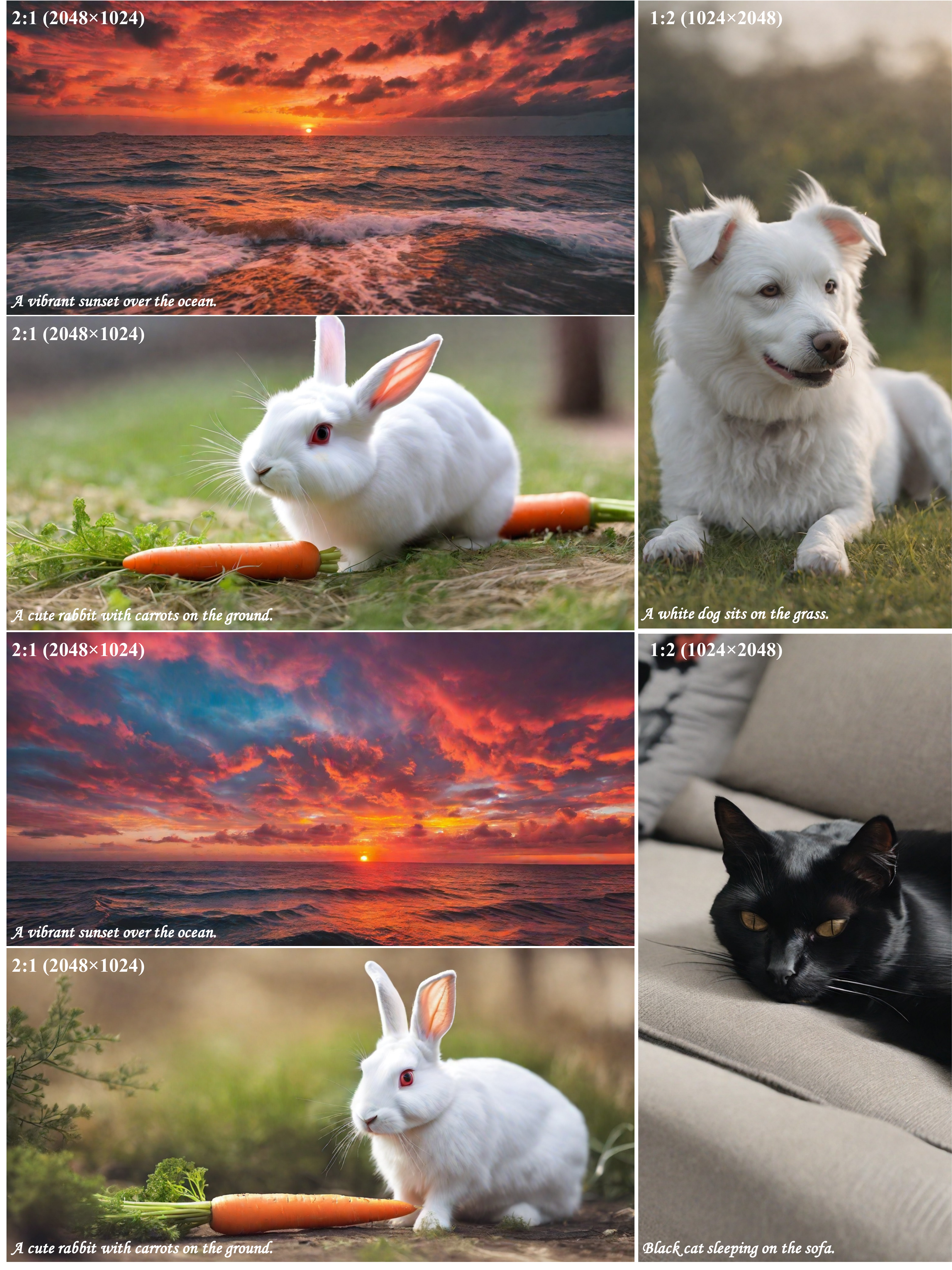} \\
  \caption{
  \textbf{More qualitative results} of applying our MegaFusion to SDXL~\cite{podell2023sdxl} model for higher-resolution image generation with various \textbf{non-standard} aspect ratios and resolutions.}
 \label{fig:SDXL_5}
\end{figure*}

\begin{figure*}[h]
  \centering
  \includegraphics[width=\textwidth]{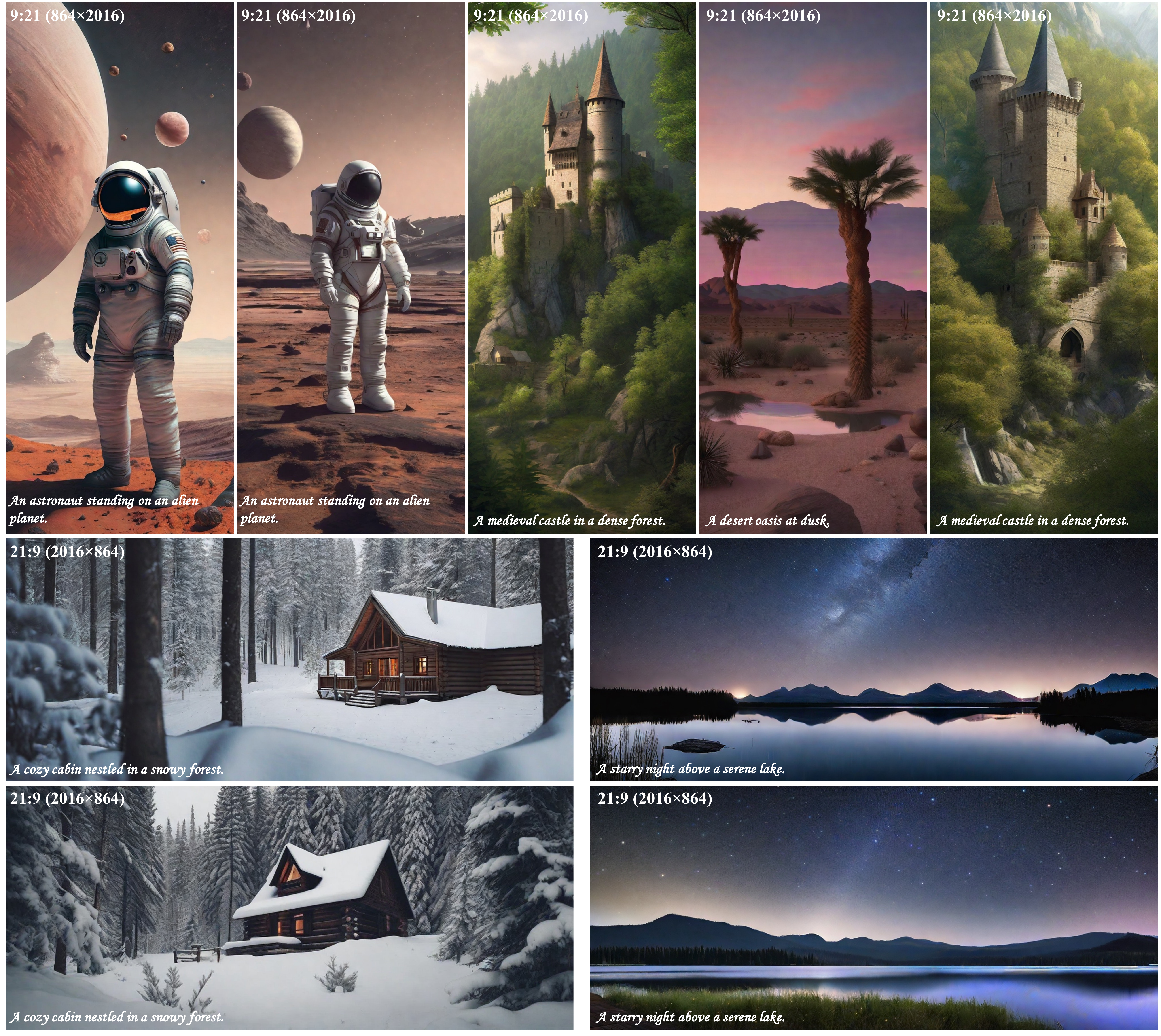} \\
  \caption{
  \textbf{More qualitative results} of applying our MegaFusion to SDXL~\cite{podell2023sdxl} model for higher-resolution image generation with various \textbf{non-standard} aspect ratios and resolutions.}
 \label{fig:SDXL_6}
\end{figure*}

\begin{figure*}[h]
  \centering
  \includegraphics[width=.9\textwidth]{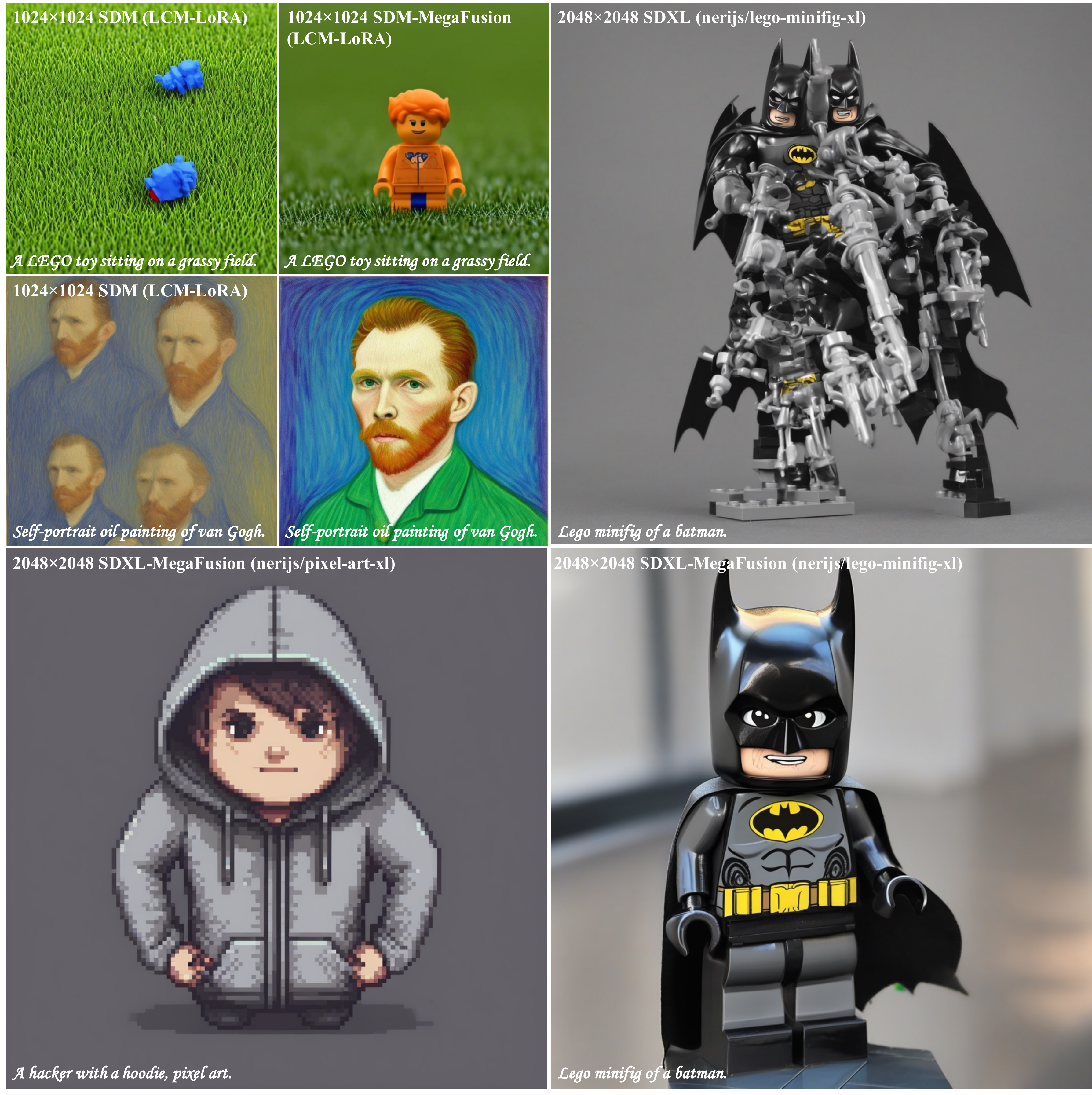} \\
  \caption{
  \textbf{Qualitative results} of applying MegaFusion to high-resolution image generation with LoRA-integrated SDM and SDXL.
  Similarly, SDM and SDXL integrated with LoRA also face common challenges like semantic deviations and object repetitions in high-resolution generation, while MegaFusion effectively addresses these challenges.
  }
 \label{fig:LoRA}
\end{figure*}

\end{document}